\PassOptionsToPackage{usenames,dvipsnames,table}{xcolor}
\documentclass[logo,copyright,12pt]{googledeepmind}

\usepackage[utf8]{inputenc}
\usepackage{verbatim}
\usepackage{amsfonts}  % Used for \mathbb.
\usepackage{amssymb}
\usepackage{physics}  % Used for \dd.
\usepackage{mathtools}  % Used for dcases environment.

\usepackage{adjustbox}
\usepackage{graphicx}
\usepackage[labelfont=bf]{caption}
\usepackage[format=hang]{subcaption}
\usepackage{pstricks, pst-node}
\usepackage{floatrow}
\usepackage{wrapfig}

\usepackage{booktabs}
\usepackage{multirow}
\usepackage{float}
\usepackage{colortbl}
\usepackage{makecell}

\usepackage[numbers,sort&compress]{natbib}
\newcommand\authoryearcite[1]{%
  (\citeauthor{#1}, \citeyear{#1})}
\newcommand\authoryearcitet[1]{%
  \citeauthor{#1} (\citeyear{#1})}

\usepackage[all]{hypcap}
\hypersetup{colorlinks}
\hypersetup{linktoc=all}
\hypersetup{citecolor=Violet}
\hypersetup{linkcolor=black}
\hypersetup{urlcolor=MidnightBlue}

\usepackage[nameinlink]{cleveref}
\creflabelformat{equation}{#1#2#3}
\Crefname{equation}{Eq.}{Eqs.}
\Crefname{assumption}{Assumption}{Assumptions}
\Crefname{prop}{Proposition}{Propositions}
\Crefname{lem}{Lemma}{Lemmas}

\usepackage{algorithm}
\usepackage[noend]{algpseudocode}

\definecolor{darkgreen}{rgb}{0,0.5,0}
\definecolor{darkred}{rgb}{0.5,0,0}

\usepackage[frozencache,cachedir=.]{minted}
\setminted{
  fontsize=\footnotesize,
  breaklines=true,
  escapeinside=|| % Specify the escape characters
}

\usepackage{amsthm}
\theoremstyle{definition}

\theoremstyle{remark}

\theoremstyle{plain}

\newtheorem{lemma}{Lemma}

\usepackage{xr}
\makeatletter
\newcommand*{\addFileDependency}[1]{
  \typeout{(#1)}
  \@addtofilelist{#1}
  \IfFileExists{#1}{}{\typeout{No file #1.}}
}
\makeatother

\usepackage{xspace}
\newcommand{\ResultsColabURL}{https://colab.research.google.com/github/google-deepmind/alphaevolve\_results/blob/master/mathematical\_results.ipynb}
\newcommand{\ResultsColab}{the accompanying \href{\ResultsColabURL}{Google Colab}}
\newcommand{\method}{\emph{AlphaEvolve}\xspace}

\usepackage[normalem]{ulem}

\usepackage{listings}

\definecolor{diffgreen}{rgb}{0.0, 0.6, 0.0}
\definecolor{diffred}{rgb}{0.8, 0.0, 0.0}
\definecolor{diffhead}{rgb}{0.5, 0.5, 0.5}
\definecolor{codegray}{rgb}{0.5,0.5,0.5}
\definecolor{codeblue}{rgb}{0.0,0.0,0.6}
\definecolor{codepurple}{rgb}{0.58,0,0.82}
\definecolor{codeorange}{rgb}{1.0, 0.5, 0.0}
\definecolor{backcolour}{rgb}{0.97,0.97,0.97}
\definecolor{highlightblue}{rgb}{0.9, 0.9, 1.0}
\definecolor{highlightyellow}{rgb}{0.95, 0.95, 0.8}
\definecolor{highlightorange}{rgb}{1.0, 0.95, 0.8}

\newcommand{\myhighlightblue}[1]{\colorbox{highlightblue}{\strut\mbox{#1}\hspace*{\fill}}}
\newcommand{\myhighlightyellow}[1]{\colorbox{highlightyellow}{\strut\mbox{#1}\hspace*{\fill}}}
\newcommand{\myhighlightorange}[1]{\colorbox{highlightorange}{\strut\mbox{#1}\hspace*{\fill}}}

\lstdefinestyle{pydiff}{
    language=Python,
    commentstyle=\color{codepurple}\itshape,
    keywordstyle=\color{codeblue}\bfseries,
    numberstyle=\tiny\color{codegray}, % Line numbers remain tiny
    stringstyle=\color{codeorange},
    basicstyle=\ttfamily\scriptsize,
    breakatwhitespace=false,
    breaklines=true,
    captionpos=b,
    keepspaces=true,
    numbers=left,
    numbersep=5pt,
    showspaces=false,
    showstringspaces=false,
    showtabs=false,
    tabsize=2,
    frame=single, % Add a frame around the code
    framerule=0.4pt,
    rulecolor=\color{lightgray},
    morecomment=[f][\color{diffgreen}]{+}, % Added lines start with +
    morecomment=[f][\color{diffred}]{-},   % Removed lines start with -
    morecomment=[f][\color{diffhead}]{@},  % Header lines start with @
    morecomment=[f][\color{codeorange}]{...},  % ... lines
    moredelim=[is][\myhighlightblue]{^1}{^}, % Use ^ as invisible marker
    moredelim=[is][\myhighlightyellow]{^2}{^}, % Use ^ as invisible marker
    moredelim=[is][\myhighlightorange]{^3}{^}, % Use ^ as invisible marker
    xleftmargin=0pt,
    xrightmargin=0pt,
}

\lstdefinestyle{pydiffmedium}{
    language=Python,
    commentstyle=\color{codepurple}\itshape,
    keywordstyle=\color{codeblue}\bfseries,
    numberstyle=\tiny\scalefont{0.69}\color{codegray}, % Line numbers remain tiny
    stringstyle=\color{codeorange},
    basicstyle=\ttfamily\scriptsize\scalefont{0.69},
    breakatwhitespace=false,
    breaklines=true,
    captionpos=b,
    keepspaces=true,
    numbers=left,
    numbersep=5pt,
    showspaces=false,
    showstringspaces=false,
    showtabs=false,
    tabsize=2,
    frame=single, % Add a frame around the code
    framerule=0.4pt,
    rulecolor=\color{lightgray},
    morecomment=[f][\color{diffgreen}]{+}, % Added lines start with +
    morecomment=[f][\color{diffred}]{-},   % Removed lines start with -
    morecomment=[f][\color{diffhead}]{@},  % Header lines start with @
    morecomment=[f][\color{codeorange}]{...},  % ... lines
    moredelim=[is][\myhighlightblue]{^1}{^}, % Use ^ as invisible marker
    moredelim=[is][\myhighlightyellow]{^2}{^}, % Use ^ as invisible marker
    moredelim=[is][\myhighlightorange]{^3}{^}, % Use ^ as invisible marker
    xleftmargin=0pt,
    xrightmargin=0pt,
}

\title{\method: A coding agent for scientific and algorithmic discovery}

\author[*]{Alexander Novikov}
\author[*]{Ngân Vũ}
\author[*]{Marvin Eisenberger}
\author[*]{Emilien Dupont}
\author[*]{Po-Sen Huang}
\author[*]{Adam Zsolt Wagner}
\author[*]{Sergey Shirobokov}
\author[*]{Borislav Kozlovskii}
\author[ \hspace{-0.2em}]{Francisco J. R. Ruiz}
\author[ \hspace{-0.2em}]{Abbas Mehrabian}
\author[ \hspace{-0.2em}]{M. Pawan Kumar}
\author[ \hspace{-0.2em}]{Abigail See}
\author[ \hspace{-0.2em}]{Swarat Chaudhuri}
\author[ \hspace{-0.2em}]{George Holland}
\author[ \hspace{-0.2em}]{Alex Davies}
\author[ \hspace{-0.2em}]{Sebastian Nowozin}
\author[ \hspace{-0.2em}]{Pushmeet Kohli}
\author[*]{Matej Balog}
\affil[ \hspace{-0.2em}]{Google DeepMind\footnote{See Acknowledgments and Author information section. $^*$Equal contributions.}}

\begin{abstract}
In this white paper, we present \method, an evolutionary coding agent that substantially enhances capabilities of state-of-the-art LLMs on highly challenging tasks such as tackling open scientific problems or optimizing critical pieces of computational infrastructure.
\method orchestrates an autonomous pipeline of LLMs, whose task is to improve an algorithm by making direct changes to the code. Using an evolutionary approach, continuously receiving feedback from one or more evaluators, \method iteratively improves the algorithm, potentially leading to new scientific and practical discoveries.
We demonstrate the broad applicability of this approach by applying it to a number of important computational problems. When applied to optimizing critical components of large-scale computational stacks at Google, \method developed a more efficient scheduling algorithm for data centers, found a functionally equivalent simplification in the circuit design of hardware accelerators, and accelerated the training of the LLM underpinning \method itself.
Furthermore, \method discovered novel, provably correct algorithms that surpass state-of-the-art solutions on a spectrum of problems in mathematics and computer science, significantly expanding the scope of prior automated discovery methods~\authoryearcite{paredes2023mathematical}.
Notably, \method developed a search algorithm that found a procedure to multiply two $4 \times 4$ complex-valued matrices using $48$ scalar multiplications; offering the first improvement, after 56 years, over Strassen's algorithm in this setting. We believe \method and coding agents like it can have a significant impact in improving solutions of  problems across many areas of science and computation.
\end{abstract}

\begin{document}

\maketitle

\section{Introduction}

\label{sec:introduction}

Discovering new high-value knowledge, such as making a novel scientific discovery or developing a commercially valuable algorithm, generally requires a prolonged process of ideation, exploration, backtracking on unpromising hypotheses, experimentation, and validation. There has been much recent interest in using large language models (LLMs) to automate significant parts of this process.
Hopes of success here are driven by the breathtaking power of recent LLMs \cite{gemini25,o3}, which can enhance their capabilities using test-time compute, and the rise of \emph{agents} that combine language generation and action \cite{yao2023react,shinn2023reflexion}. These advances have improved performance across a range of established benchmarks and accelerated discovery-oriented tasks like hypothesis generation~\cite{gottweis2025towards} and experiment design \cite{huang2024crisprgpt,boiko2023autonomous}. However, getting LLM pipelines all the way to making entirely new scientific or practical discoveries remains challenging. 

In this white paper, we present an LLM code superoptimization agent, called \method, that takes on this challenge using a combination of evolutionary computation and LLM-based code generation. \method focuses on the broad spectrum of scientific and engineering discovery problems in which the candidates of discovery can be automatically evaluated. 
It represents the candidates (for example, new mathematical objects or practical heuristics) as algorithms and uses a set of LLMs to generate, critique, and evolve a pool of such algorithms. The LLM-directed evolution process is grounded using code execution and automatic evaluation. 
This evaluation mechanism allows \method to avoid any incorrect suggestions from the base LLM~\citep{huang2025survey}.

The evolutionary process in \method leverages modern LLMs' ability to respond to feedback, enabling the discovery of candidates that are substantially different from the initial candidate pool in syntax and function.
It is applicable both to problems where discovering new algorithms is the intrinsic goal, as well as to the broad range of problems where the solution of interest is not an algorithm itself but an algorithm can \emph{describe} how that solution is to be constructed or found.
In the latter case, discovering the algorithm is only an instrumental goal, but it turns out to be a surprisingly effective strategy compared to searching for the solution directly~\cite{paredes2023mathematical}.

The idea of combining evolutionary methods with coding LLMs has been previously explored in various specialized settings.
In particular, \method is a substantial enhancement of \emph{FunSearch}~\cite{paredes2023mathematical} (see~\Cref{tab:funsearch-vs-alphaevolve}), which used LLM-guided evolution to discover heuristics in order to construct novel mathematical objects or to drive the operation of online algorithms.
Also, related approaches have been used in tasks such as discovering policies for simulated robots~\cite{lehman2023evolution}, symbolic regression \citep{shojaee2025llmsr,grayeli2024symbolic}, and the synthesis of heuristic functions for combinatorial optimization~\cite{liu2024evolution}.
In contrast to these systems, \method leverages state-of-the-art (SOTA) LLMs to evolve large pieces of code that implement complex algorithms spanning multiple functions and components. As a result, it is able to go significantly beyond its predecessors in scale and generality.

\begin{table}[H]
\small
\rowcolors{2}{white}{gray!20}
\begin{center}
\begin{tabular}{ll} \toprule
    \emph{FunSearch}~\cite{paredes2023mathematical} & \method \\
    \midrule
    evolves single function & evolves entire code file\\
    evolves up to 10-20 lines of code & evolves up to hundreds of lines of code\\
    evolves code in Python & evolves any language\\
    needs fast evaluation ($\leq 20$min on 1 CPU)\;\; & can evaluate for hours, in parallel, on accelerators\\
    millions of LLM samples used & thousands of LLM samples suffice\\
    small LLMs used; no benefit from larger & benefits from SOTA LLMs\\ 
    minimal context (only previous solutions) & rich context and feedback in prompts\\
    optimizes single metric & can simultaneously optimize multiple metrics\\
    \bottomrule
\end{tabular}
\caption{Capabilities and typical behaviours of \method and our previous agent.}
\label{tab:funsearch-vs-alphaevolve}
\end{center}
\end{table}

While the use of an automated evaluation metric offers \method a key advantage, it is also a limitation---in particular, it puts tasks that require manual experimentation out of our scope. 
Because problems in mathematics, computer science, and system optimization typically permit automated evaluation metrics, our efforts on \method focus on these domains.
Specifically, we use \method to make progress on several well-known open problems in algorithm design and constructive mathematics, as well as the optimization of critical layers in the large-scale computation stacks at Google.

Within algorithm design, we consider the fundamental problem of discovering fast algorithms for multiplying matrices, a problem to which a more specialized AI approach had been applied previously~\citep{fawzi2022discovering}.
Despite being general-purpose, \method goes beyond~\cite{fawzi2022discovering}, improving the SOTA for 14 matrix multiplication algorithms; notably, for $4 \times 4$ matrices, \method improves \authoryearcitet{strassen1969gaussian}'s algorithm by discovering an algorithm using 48 multiplications to multiply $4 \times 4$ complex-valued matrices.%
\footnote{These discovered algorithms as well as our other new mathematical results can be found at \url{\ResultsColabURL}.}

In mathematics, we consider a broad range of open problems on which one can make progress by discovering constructions (objects) with better properties than all previously known constructions, according to given mathematical definitions.
We apply \method to a large number (over 50) of such problems and match the best known constructions on $\sim$75\% of them (in many cases these constructions are likely to already be optimal). On $\sim$20\% of the problems, \method surpasses the SOTA and discovers new, provably  better constructions.
This includes an improvement on the Minimum Overlap Problem set by \citet{erdHos1955some} and an improved construction on the Kissing Numbers problem in $11$ dimensions~\citep{kissing_survey,kissing_11}.

Finally, we use \method in four engineering problems spanning different layers of Google's compute stack: discovering scheduling heuristics for Google's cluster management system, optimizing matrix-multiplication kernels used to train LLMs, optimizing arithmetic circuits used within TPUs, and optimizing the runtime of attention in Transformers.
Because these components are run repeatedly over a long period of time, any improvements are highly valuable.

\section{\method}
\label{sec:method}

\begin{wrapfigure}{r}{0.5\textwidth}
\vspace{-4.5em}
\includegraphics[width=0.99\linewidth, trim=2cm 2cm 2cm 2cm]{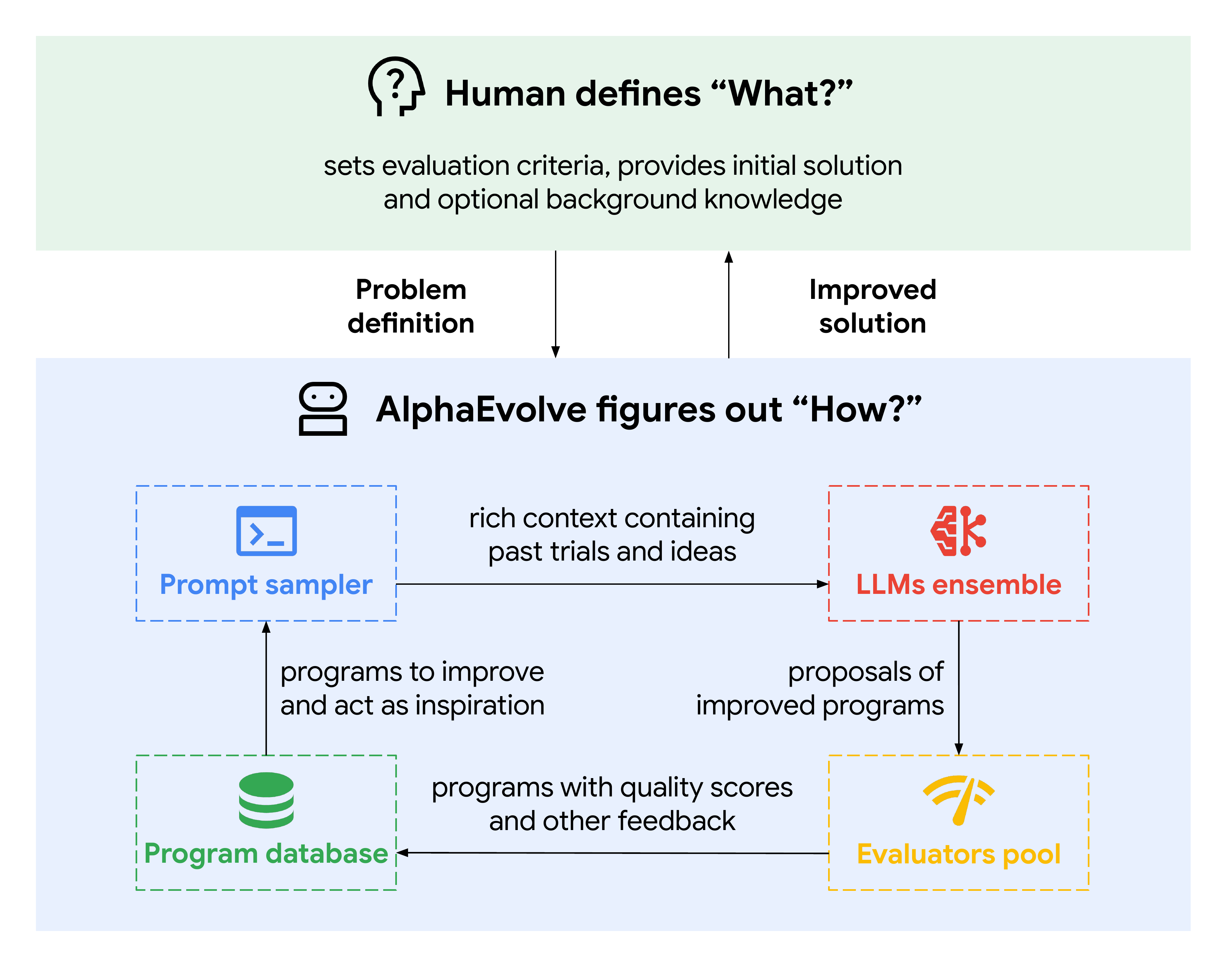}
\caption{\method high-level overview.}
\label{fig:high-level}
\end{wrapfigure}

\begin{figure}[tb]
    \centering
    \includegraphics[width=0.96\textwidth, trim=0cm 0cm 0cm 0cm, clip]{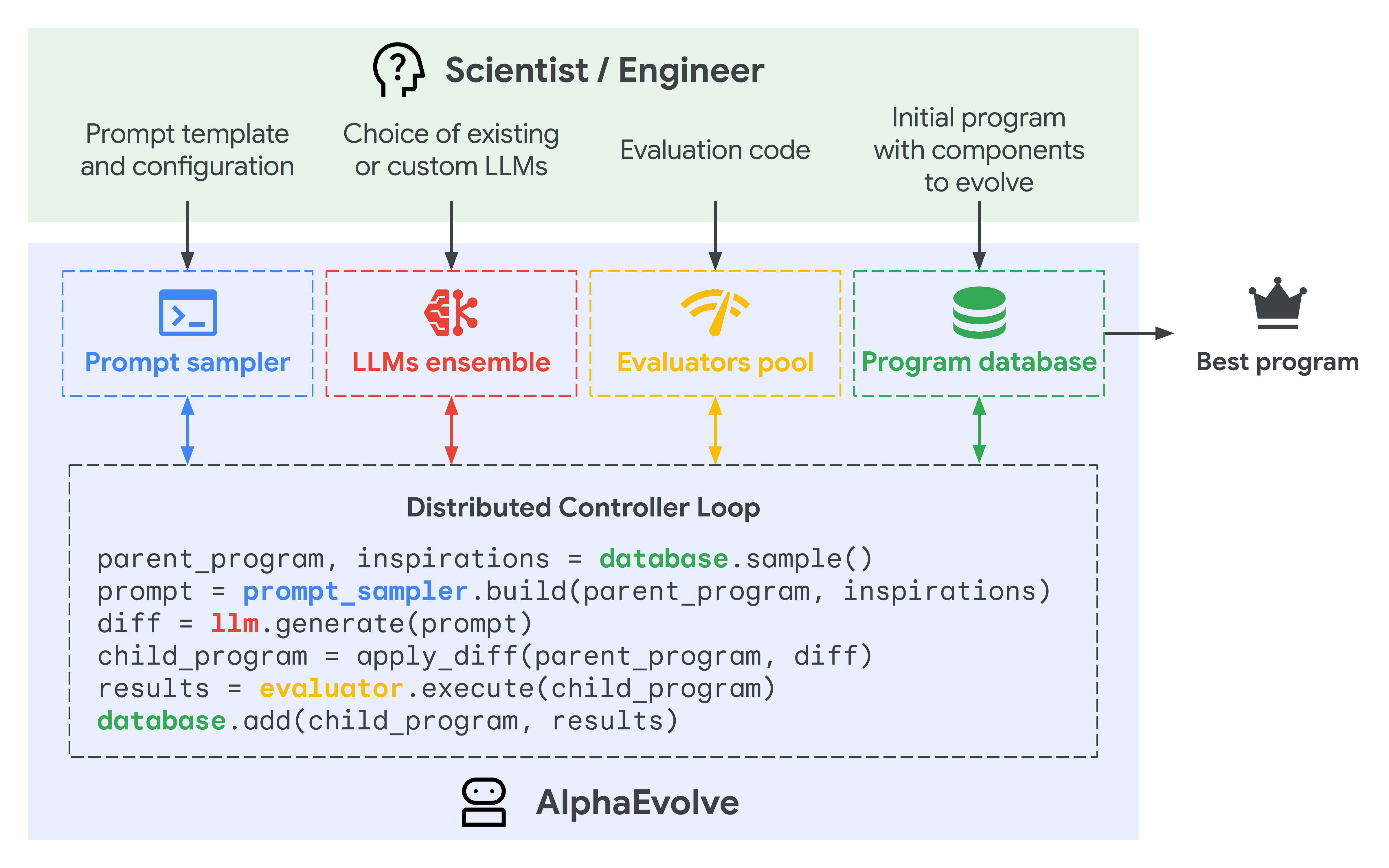}
    \caption{%
    Expanded view of the \method discovery process. The user provides an initial program (with components to evolve marked), evaluation code, and optional configurations (\Cref{subsec:specification}).
    \method then initiates an evolutionary loop.
    The \emph{Prompt sampler} uses programs from the \emph{Program database} to construct rich prompts (\Cref{subsec:prompting}).
    Given these prompts, the \emph{LLMs} generate code modifications (diffs), which are applied to create new programs (\Cref{subsec:generation}).
    These are then scored by \emph{Evaluators} (\Cref{subsec:evaluation}), and promising solutions are registered back into the \emph{Program database} (\Cref{subsec:evolution}), driving the iterative discovery of better and better programs.
    }
    \label{fig:method}
\end{figure}

\method is a coding agent that orchestrates an autonomous pipeline of computations including queries to LLMs, and produces algorithms that address a user-specified task.
At a high level, the orchestrating procedure is an evolutionary algorithm that gradually develops programs that improve the score on the automated evaluation metrics associated with the task.
A high-level overview of \method is shown in~\Cref{fig:high-level}, and~\Cref{fig:method} gives an expanded view.

\subsection{Task specification}
\label{subsec:specification}

\paragraph{Evaluation.} Since \method tackles problems with machine-gradeable solutions, the user must provide a mechanism for automatically assessing generated solutions.
This mechanism takes the form of a function $h$ mapping a solution to a set of scalar evaluation metrics.
By convention, these metrics are maximized.
In our current setup, $h$ is typically implemented as a Python function, called \texttt{evaluate}, with a fixed input/output signature, returning a dictionary of scalars.

Depending on the application, executing this function may take only seconds on a single device or spawn extensive computations. For mathematical problems, the function $h$ is typically very simple.
For example, when wishing to find largest possible graphs satisfying a given property, $h$ invokes the evolved code to generate a graph, checks whether the property holds, and then simply returns the size of the graph as the score.
In more complicated cases, the function $h$ might involve performing an evolved search algorithm, or training and evaluating a machine learning model.

\begin{figure}[htbp]
    \centering
    \begin{minipage}[c]{0.48\textwidth} %
        \centering
        \begin{subfigure}[t]{\textwidth} %
            \centering
\begin{minted}[fontsize=\tiny, bgcolor=backcolour]{python}
# EVOLVE-BLOCK START
"""Image classification experiment in jaxline."""

import jax
...
# EVOLVE-BLOCK-END

...

# EVOLVE-BLOCK-START
class ConvNet(hk.Module):
  def __init__(self, num_classes): ...
  def __call__(self, inputs, is_training): ...


def sweep():
  return hyper.zipit([...])
# EVOLVE-BLOCK-END

...

def evaluate(eval_inputs) -> dict[str, float]:
  ...
  return metrics
\end{minted}
            \captionsetup{justification=centering, singlelinecheck=false}
            \caption{\raisebox{-0.3\baselineskip}{\includegraphics[height=1.8em]{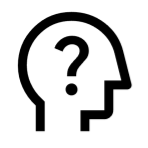}}}
            \label{fig:grounding-api}
        \end{subfigure}
        \vfill %
        \begin{subfigure}[b]{\textwidth} %
            \centering

\begin{minted}[fontsize=\tiny, bgcolor=backcolour, escapeinside=||, breaksymbolleft={}]{python}
|The current model uses a simple ResNet architecture with only three ResNet blocks. We can improve its performance by increasing the model capacity and adding regularization. This will allow the model to learn more complex features and generalize better to unseen data. We also add weight decay to the optimizer to further regularize the model and prevent overfitting. AdamW is generally a better choice than Adam, especially with weight decay.|


<<<<<<< SEARCH
    self._block1 = ResNetBlock(num_channels)
    self._block2 = ResNetBlock(num_channels * 2, stride=2)
    self._block3 = ResNetBlock(num_channels * 4, stride=2)
=======
    self._block1 = ResNetBlock(num_channels)
    self._block2 = ResNetBlock(num_channels, stride=1)
    self._block3 = ResNetBlock(num_channels * 2, stride=2)
    self._block4 = ResNetBlock(num_channels * 2, stride=1)
    self._block5 = ResNetBlock(num_channels * 4, stride=2)
    self._block6 = ResNetBlock(num_channels * 4, stride=1)
>>>>>>> REPLACE
<<<<<<< SEARCH
  def optimizer(self, learning_rate):
    return optax.adam(learning_rate)
=======
  def optimizer(self, learning_rate):
    return optax.adamw(learning_rate, weight_decay=1e-4)
>>>>>>> REPLACE
\end{minted}
            \addtocounter{subfigure}{+1}
            \captionsetup{justification=centering, singlelinecheck=false}
            \caption{\raisebox{-0.5\baselineskip}{\includegraphics[height=2.0em]{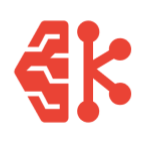}}}
            \label{fig:grounding-llm}
        \end{subfigure}
    \end{minipage}
    \hfill %
    \begin{minipage}[c]{0.48\textwidth} %
        \centering
        \begin{subfigure}{\textwidth} %
            \centering

\begin{minted}[fontsize=\tiny, bgcolor=backcolour, escapeinside=||, breaksymbolleft={}]{python}
|Act as an expert software developer. Your task is to iteratively improve the provided codebase. [...]

- Prior programs

Previously we found that the following programs performed well on the task at hand:|

top_1_acc: 0.796; neg_eval_log_loss: 0.230; average_score: 0.513

"""Image classification experiment in jaxline."""
[...]
class ConvNet(hk.Module):
  """Network."""

  def __init__(self, num_channels=32, num_output_classess=10):
    super().__init__()
    self._conv1 = hk.Conv2D(num_channels, kernel_shape=3)
    self._conv2 = hk.Conv2D(num_channels * 2, kernel_shape=3)
    self._conv3 = hk.Conv2D(num_channels * 4, kernel_shape=3)
    self._logits_module = hk.Linear(num_output_classes)
[...]


|- Current program

Here is the current program we are trying to improve (you will need to propose a modification to it below).|

top_1_acc: 0.862; neg_eval_log_loss: 0.387; average_score: 0.624

"""Image classification experiment in jaxline."""
[...]
class ConvNet(hk.Module):
  """Network."""

  def __init__(self, num_channels=32, num_output_classes=10):
    super().__init__()
    self._conv1 = hk.Conv2D(num_channels, kernel_shape=3)
    self._block1 = ResNetBlock(num_channels)
    self._block2 = ResNetBlock(num_channels * 2, stride=2)
    self._block3 = ResNetBlock(num_channels * 4, stride=2)
    self._logits_module = hk.Linear(num_output_classes)
|[...]

SEARCH/REPLACE block rules:
[...]

Make sure that the changes you propose are consistent with each other. For example, if you refer to a new config variable somewhere, you should also propose a change to add that variable.

Example:
[...]

Task
Suggest a new idea to improve the code that is inspired by your expert knowledge of optimization and machine learning.  

Describe each change with a SEARCH/REPLACE block.|
\end{minted}
            \addtocounter{subfigure}{-2}
            \captionsetup{justification=centering, singlelinecheck=false}
            \caption{\raisebox{-0.5\baselineskip}{\includegraphics[height=2.0em]{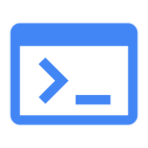}}}
            \label{fig:grounding-prompt}
        \end{subfigure}
    \end{minipage}
    \caption{Illustrative example of applying \method to evolving a supervised learning pipeline. All snippets are abbreviated, with ellipsis (...) indicating skipped lines. (a) The user-provided file with blocks marked for evolution, and the special \texttt{evaluate} function that can be invoked to score the current version of the code. (b) Example of an assembled prompt to be provided to the LLMs. (c) Example output generated by the LLM. The proposed diffs in (c) will be applied to the "current program" shown in the prompt (b), and the resulting modified program will then be sent to the evaluators. The evaluators will invoke the \texttt{evaluate} function from (a) in order to obtain the scores of the newly proposed program.}
    \label{fig:grounding}
\end{figure}

\paragraph{API.} To support evolving multiple components across a codebase, \method exposes an input API where blocks of code can be annotated as to-be-evolved-by-the-system; see Figure~\ref{fig:grounding-api} for an illustration. This design facilitates integrating it with existing codebases while requiring only minimal changes, simply by adding special markers (\texttt{\# EVOLVE-BLOCK-START} and \texttt{\# EVOLVE-BLOCK-END}) as comments into the code.

Any user-provided code inside such evolution blocks serves as the initial solution to be improved by \method, and the rest of the code forms a skeleton that ties the evolved pieces together, so that they can be invoked from \texttt{evaluate}. 
While this initial implementation must be complete, it can be rudimentary---for instance, consisting of single-line functions that return constants of the appropriate types.

\paragraph{Flexibility in choosing the abstraction.} \method can be applied to the same problem in very different ways---especially when the evolved programs are not the final output but a means to discover solutions.
For example, \method can evolve the solution in raw string representation (as in classical evolutionary algorithms); evolve a function of a definite form that specifies how to construct the solution from scratch (the approach taken in~\cite{paredes2023mathematical}); evolve a bespoke search algorithm to find the solution within some fixed compute budget; or even co-evolve intermediate solutions and search algorithms together, such that each search algorithm is specifically tailored to further improve upon a particular intermediate solution.

We find that different levels of abstraction work better for different problems.
For example, we hypothesize that for problems with highly symmetric solutions it is advantageous to evolve constructor functions as these tend to be more concise~\cite{paredes2023mathematical}, whereas for problems with non-symmetric solutions it works better to evolve customized search algorithms.

\subsection{Prompt sampling}
\label{subsec:prompting}

As \method leverages SOTA LLMs, it supports various types of customization and providing long contexts as part of the primary evolution prompt.
This prompt comprises multiple previously discovered solutions sampled from the program database, as well as system instructions on how to propose changes to a particular solution.
Beyond these key ingredients, users can further tailor prompts to their specific needs in different ways, such as the following.
\begin{itemize}
\item
\emph{Explicit context}: details about the problem being solved, such as fixed human-written instructions, equations, code snippets, or relevant literature (e.g., pdf files).
\item
\emph{Stochastic formatting}: template placeholders with human-provided alternatives for increased diversity, instantiated using probability distributions provided in a separate config file.
\item
\emph{Rendered evaluation results}: usually this will include a program, the result of executing that program, and the scores assigned by the \texttt{evaluate} function.
\item
\emph{Meta prompt evolution}: instructions and context suggested by the LLM itself in an additional prompt-generation step, co-evolved in a separate database analogous to the solution programs.
\end{itemize}

\subsection{Creative generation}
\label{subsec:generation}

To drive the evolutionary procedure, \method leverages the capabilities of SOTA LLMs, whose principal role is to digest information about previously developed solutions and propose new, diverse ways to improve the solutions.
Although \method is model-agnostic, in ablations we observe that \method performs increasingly better as the underlying LLM improves (see~\Cref{sec:ablations_rewrite}).

\paragraph{Output format.}
When \method asks an LLM to modify existing code, especially within larger codebases, it requests the changes to be provided as a sequence of diff blocks in a specific format:

\begin{verbatim}
<<<<<<< SEARCH
  # Original code block to be found and replaced
=======
  # New code block to replace the original
>>>>>>> REPLACE
\end{verbatim}

Here, the code between \texttt{<{}<{}<{}<{}<{}<{}< SEARCH} and \texttt{=======} is the exact segment to match in the current program version. The code between \texttt{=======} and \texttt{>{}>{}>{}>{}>{}>{}> REPLACE} is the new segment that will replace the original one.
This allows for targeted updates to specific parts of the code.

In cases where the code being evolved is very short, or when a complete rewrite is more appropriate than a small modification, \method can be configured to instruct the LLM to output the entire code block directly, rather than using the diff format. 

\paragraph{Models used.}
\method employs an ensemble of large language models. Specifically, we utilize a combination of Gemini 2.0 Flash and Gemini 2.0 Pro.
This ensemble approach allows us to balance computational throughput with the quality of generated solutions.
Gemini 2.0 Flash, with its lower latency, enables a higher rate of candidate generation, increasing the number of ideas explored per unit of time.
Concurrently, Gemini 2.0 Pro, possessing greater capabilities, provides occasional, higher-quality suggestions that can significantly advance the evolutionary search and potentially lead to breakthroughs.
This strategic mix optimizes the overall discovery process by maximizing the volume of evaluated ideas while retaining the potential for substantial improvements driven by the more powerful model.

\subsection{Evaluation}
\label{subsec:evaluation}

To track \method's progress and to select which ideas to propagate in future generations, each new solution proposed by the LLMs is automatically evaluated.
In principle, this process amounts to simply executing the user-provided evaluation function $h$ on the generated solution.
In practice, \method supports optional mechanisms to make this evaluation more flexible and more efficient:
\begin{itemize}
\item
\emph{Evaluation cascade (hypothesis testing)}: the user can specify ensembles of test cases of increasing difficulty, such that new solutions are  evaluated on the next stage only if they achieve sufficiently promising results in all earlier stages.
This helps to prune out less promising solutions more quickly. Moreover, new solutions are initially evaluated on a small scale before being subjected to the main test cases, to filter out faulty programs early.
\item
\emph{LLM-generated feedback}: in some applications, desirable solutions have certain characteristics that are difficult to capture precisely in the user-provided evaluation function $h$; for example, simplicity of the discovered program.
These properties can be graded using separate LLM calls and added to the dictionary of scores to steer evolution, or they can be used to discard solutions when a criterion is not fulfilled.
\item
\emph{Parallelized evaluation}: the sample efficiency of \method makes it feasible to spend on the order of 100 compute-hours to evaluate any new solution.
However, unless individual evaluations are parallelized to reduce their wall-clock duration, this can slow down the rate at which new generations appear, limiting the ability of the evolutionary algorithm to apply several consecutive mutations.
In many applications, evaluation is embarrassingly parallel (for example, running a search algorithm from multiple randomized initializations), allowing \method to distribute this work through asynchronous calls to an evaluation cluster.
\end{itemize}

\paragraph{Multiple scores.}
\method allows for optimizing multiple user-provided scores, i.e., evolving objects that achieve a high score under one or multiple evaluation metrics.
This has both an intrinsic and instrumental value. While in multiple applications we genuinely care about developing solutions for multiple evaluation metrics (or one solution that is strong on all of them simultaneously), we find that even if one metric is of particular interest, optimizing for multiple metrics often improves results for the single target metric.
Perhaps this occurs because programs excelling under different evaluation criteria often possess distinct structures or logic and, by incorporating examples of these diverse, high-performing programs---each representing a different definition of ``good''---into the prompts provided to the language model, we can stimulate the generation of more varied candidate solutions, increasing the chances of discovering novel approaches that are highly effective for the target metric.

\subsection{Evolution}
\label{subsec:evolution}

During its evolutionary procedure, \method continually generates a growing number of solutions with evaluation results (scores and program outputs) attached to them.
These solutions are stored in an evolutionary database, the primary goal of which is to optimally resurface previously explored ideas in future generations.
A key challenge in designing such databases is balancing exploration and exploitation, to continuously improve the best programs while maintaining diversity to encourage exploration of the entire search space.
In \method, the evolutionary database implements an algorithm that is inspired by a combination of the MAP elites algorithm~\cite{mouret2015illuminating} and island-based population models~\cite{tanese1989distributed, paredes2023mathematical}.

\subsection{Distributed pipeline}
\label{subsec:pipeline}

\method is implemented as an asynchronous computational pipeline (using the \texttt{asyncio} Python library) in which many computations are run concurrently, with each computation blocking (waiting) whenever its next step relies on the result of another, yet unfinished computation.
More specifically, the asynchronous pipeline comprises a controller, LLM samplers, and evaluation nodes.
The entire pipeline is optimized for throughput (rather than the speed of any one particular computation), in order to maximize the number of ideas that can be proposed and evaluated within a specific overall computation budget.

\section{Results}
\label{sec:results}

\subsection{Faster matrix multiplication via finding novel algorithms for tensor decomposition}
\label{subsec:matmul}

\begin{table}[t]
\rowcolors{2}{white}{lightgray}
\begin{center}
    \begin{tabular}{ccc}
    $\langle m, n, p \rangle$ & best known [reference] & \method \\ \midrule
    $\langle 2, 4, 5 \rangle$ & 33 \citep{hopcroft} & \textbf{32} \\
    $\langle 2, 4, 7 \rangle$ & 46 \citep{smirnov2013bilinear} & \textbf{45} \\
    $\langle 2, 4, 8 \rangle$ & 52 \citep{smirnov2013bilinear} & \textbf{51} \\
    $\langle 2, 5, 6 \rangle$ & 48 \citep{smirnov2013bilinear}   & \textbf{47}    \\ 
    $\langle 3, 3, 3 \rangle$ & 23 \citep{laderman}   & 23   \\
    $\langle 3, 4, 6 \rangle$ & 56 \citep{Kauers_2025}   & \textbf{54}   \\ 
    $\langle 3, 4, 7 \rangle$ & 66 \citep{smirnov2021}   & \textbf{63}   \\ 
    $\langle 3, 4, 8 \rangle$ & 75 \citep{smirnov2021}   & \textbf{74}    \\ 
    $\langle 3, 5, 6 \rangle$ & 70 \citep{Kauers_2025}   & \textbf{68}    \\
    $\langle 3, 5, 7 \rangle$ & 82 \citep{smirnov2021}   & \textbf{80}    \\ 
    $\langle 4, 4, 4 \rangle$ & 49 \citep{strassen1969gaussian}   & \textbf{48}  \\ 
    $\langle 4, 4, 5 \rangle$ & 62 \citep{kauers2023flip}   & \textbf{61}    \\
    $\langle 4, 4, 7 \rangle$ & 87 \citep{smirnov2013bilinear} & \textbf{85}    \\
    $\langle 4, 4, 8 \rangle$ & 98~\citep{strassen1969gaussian} & {\textbf{96}} \\ 
    $\langle 4, 5, 6 \rangle$ & 93 \citep{Kauers_2025}   & \textbf{90}    \\ 
    $\langle 5, 5, 5 \rangle$ & 93 \citep{flip_graphs_with_symmetry}  & 93  \\ 
    \end{tabular}
\caption{%
Upper bounds on the rank of the tensor $\langle m,n,p \rangle$ representing the product of an $m\times n$ matrix and an $n\times p$ matrix, i.e.~the number of scalar multiplications required to compute this matrix product.
Beyond the examples shown here, for all parameters $m,n,p\leq 5$, \method either matched or surpassed the best known solutions, and provided exact algorithms (see \Cref{tab:relaxed-opt-results-appendix} in appendix for full results).
For $\langle 3, 4, 7\rangle$, $\langle 4, 4, 4\rangle$, and $\langle 4, 4, 8\rangle$, the algorithms discovered by \method use complex-valued multiplications which can be used for exact multiplication of complex or real-valued matrices.
The decompositions shown in this table can be found in \ResultsColab.%
}
    \label{tab:relaxed-opt-results}
\end{center}
\end{table}

From accelerating machine learning computations to enabling realistic computer graphics, matrix multiplication serves as a fundamental operation underpinning numerous critical algorithms and applications within computer science.
Since the pioneering work of \citet{strassen1969gaussian}, it has been known that a rich space of algorithms for multiplying two matrices can be represented as decompositions of a given 3D tensor into rank-one tensors.
The rank (number of terms) of the decomposition exactly specifies the number of scalar multiplications needed to compute the matrix product.
Hence, to develop faster matrix multiplication algorithms one needs to find low-rank decompositions of particular tensors.
This problem has been tackled with many approaches, from specialized alternating least squares solvers~\citep{smirnov2013bilinear} to deep reinforcement learning~\citep{fawzi2022discovering} and custom search algorithms~\cite{kauers2023flip}; yet, despite decades of effort, even for the simple case of multiplying two $3\times 3$ matrices, the minimum achievable rank is not known, showcasing the difficulty of the problem.

Starting from the problem description and a standard gradient-based algorithm (including an initializer, a reconstruction loss function, and an Adam optimizer~\cite{kingma2015adam}), \method is able to develop sophisticated tensor decomposition algorithms that outperform existing approaches.
To evaluate each evolved program, we choose a set of matrix multiplication targets and run the algorithm, initialized with multiple random seeds using the evaluation cascade described in \Cref{subsec:evaluation}. 
The performance is then measured as the best (lowest) rank achieved on each target as well as the fraction of seeds that achieved this rank, providing a signal for \method to hill-climb.
To ensure the exactness of the decomposition and avoid any potential numerical error, when evaluating, we round each element to the nearest integer or the nearest half-integer; and, to encourage the algorithm to generate near-integral solutions, we include this request in natural language in the LLM's prompt.

\begin{figure}[p]
\vspace{-0.03\textwidth}
\hspace{-0.1\textwidth}
\begin{minipage}{0.32\textwidth}
% Does not work in newer versions of pdflatex -- runs out of memory.
%\lstinputlisting[style=pydiffsmall, backgroundcolor=\color{backcolour}]{code/diff_full.diff}
\includegraphics[width=\linewidth]{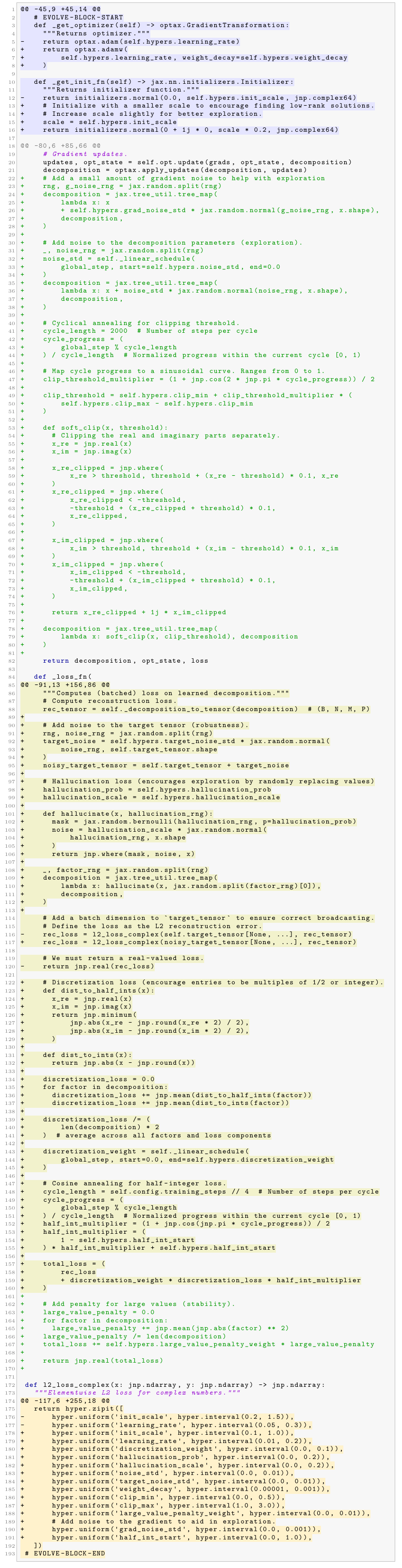}
\end{minipage}
\hspace{0.06\textwidth}
\begin{minipage}{0.59\textwidth}%
\begin{lstlisting}[style=pydiffmedium, backgroundcolor=\color{highlightblue}]
@@ -45,9 +45,14 @@
   # EVOLVE-BLOCK-START
   def _get_optimizer(self) -> optax.GradientTransformation:
     """Returns optimizer."""
-    return optax.adam(self.hypers.learning_rate)
+    return optax.adamw(
+        self.hypers.learning_rate, weight_decay=self.hypers.weight_decay
+    )
 
   def _get_init_fn(self) -> jax.nn.initializers.Initializer:
     """Returns initializer function."""
-    return initializers.normal(0.0, self.hypers.init_scale, jnp.complex64)
+    # Initialize with a smaller scale to encourage finding low-rank solutions.
+    # Increase scale slightly for better exploration.
+    scale = self.hypers.init_scale
+    return initializers.normal(0 + 1j * 0, scale * 0.2, jnp.complex64)
\end{lstlisting}
\vspace{-0.4em}
\begin{lstlisting}[style=pydiffmedium, backgroundcolor=\color{highlightyellow}]
@@ -91,13 +156,86 @@
     """Computes (batched) loss on learned decomposition."""
     # Compute reconstruction loss.
     rec_tensor = self._decomposition_to_tensor(decomposition)  # (B, N, M, P)
...
+    # Discretization loss (encourage entries to be multiples of 1/2 or integer).
+    def dist_to_half_ints(x):
...
+
+    def dist_to_ints(x):
...
+    discretization_loss = 0.0
+    for factor in decomposition:
+      discretization_loss += jnp.mean(dist_to_half_ints(factor))
+      discretization_loss += jnp.mean(dist_to_ints(factor))
+
+    discretization_loss /= (
+        len(decomposition) * 2
+    )  # average across all factors and loss components
+
+    discretization_weight = self._linear_schedule(
+        global_step, start=0.0, end=self.hypers.discretization_weight
+    )
+
+    # Cosine annealing for half-integer loss.
+    cycle_length = self.config.training_steps // 4  # Number of steps per cycle
+    cycle_progress = (
+        global_step % cycle_length
+    ) / cycle_length  # Normalized progress within the current cycle [0, 1)
+    half_int_multiplier = (1 + jnp.cos(jnp.pi * cycle_progress)) / 2
+    half_int_multiplier = (
+        1 - self.hypers.half_int_start
+    ) * half_int_multiplier + self.hypers.half_int_start
+
+    total_loss = (
+        rec_loss
+        + discretization_weight * discretization_loss * half_int_multiplier
+    )
...
\end{lstlisting}
\vspace{-0.4em}
\begin{lstlisting}[style=pydiffmedium, backgroundcolor=\color{highlightorange}]
@@ -117,6 +255,18 @@
   return hyper.zipit([
-      hyper.uniform('init_scale', hyper.interval(0.2, 1.5)),
-      hyper.uniform('learning_rate', hyper.interval(0.05, 0.3)),
+      hyper.uniform('init_scale', hyper.interval(0.1, 1.0)),
+      hyper.uniform('learning_rate', hyper.interval(0.01, 0.2)),
+      hyper.uniform('discretization_weight', hyper.interval(0.0, 0.1)),
+      hyper.uniform('hallucination_prob', hyper.interval(0.0, 0.2)),
+      hyper.uniform('hallucination_scale', hyper.interval(0.0, 0.2)),
+      hyper.uniform('noise_std', hyper.interval(0.0, 0.01)),
+      hyper.uniform('target_noise_std', hyper.interval(0.0, 0.01)),
+      hyper.uniform('weight_decay', hyper.interval(0.00001, 0.001)),
+      hyper.uniform('clip_min', hyper.interval(0.0, 0.5)),
+      hyper.uniform('clip_max', hyper.interval(1.0, 3.0)),
+      hyper.uniform('large_value_penalty_weight', hyper.interval(0.0, 0.01)),
+      # Add noise to the gradient to aid in exploration.
+      hyper.uniform('grad_noise_std', hyper.interval(0.0, 0.001)),
+      hyper.uniform('half_int_start', hyper.interval(0.0, 1.0)),
   ])
 # EVOLVE-BLOCK-END
\end{lstlisting}
\end{minipage}
\hspace{-0.1\textwidth}
\caption{Changes proposed by \method to discover faster matrix multiplication algorithms. The full diff is outlined on the left (see magnified version in~\Cref{fig:relaxed-opt-diff-appendix-1,fig:relaxed-opt-diff-appendix-2,fig:relaxed-opt-diff-appendix-3}) and some excerpts are highlighted on the right. In this example, \method proposes extensive changes across several components, including the optimizer and weight initialization (top right), the loss function (middle right), and hyperparameter sweep (bottom right). These changes are highly non-trivial, requiring 15 mutations during the evolutionary process.}\label{fig:relaxed-opt-diff}
\centering
\end{figure}

In Table~\ref{tab:relaxed-opt-results}, one can see that the various algorithms developed by \method improve the state of the art for 14 different matrix multiplication targets.
Notably, for multiplying two $4\times 4$ matrices, applying the algorithm of \citet{strassen1969gaussian} recursively results in an algorithm with rank (number of scalar multiplications) equal to 49, which works over any field.
For the very specific case of multiplying in the field with 2 elements,~\citet{fawzi2022discovering} found an algorithm with rank 47.
For 56 years, designing an algorithm with rank less than 49 over any field with characteristic 0 was an open problem.%
\footnote{There exist algorithms using fewer than 49 multiplications, but they do not correspond to decompositions of the matrix multiplication tensor, and they cannot be applied recursively to multiplying larger matrices.\vspace{-1em}}
\method is the first method to find a rank-$48$ algorithm to multiply two $4\times 4$ complex-valued matrices.

As shown in Figure~\ref{fig:relaxed-opt-diff}, \method makes significant changes to the initial program, introducing several original ideas to design increasingly better algorithms.
While most results in Table~\ref{tab:relaxed-opt-results} (including $\langle 4, 4, 4 \rangle$) were obtained from a simple initial program, we found that for some parameters, seeding the initial program with our own ideas (such as adding stochasticity to the evaluation function or using evolutionary approaches) could further boost performance, highlighting the possibility of scientific collaboration between researchers and \method.

\subsection{Finding tailored search algorithms for a wide range of open mathematical problems}
\label{subsec:math}

A significant frontier in mathematical research involves discovering objects or \textit{constructions} that possess optimal, or near-optimal, properties according to some measure. Examples range from finding dense packings of geometric shapes~\cite{geometry_collection} to identifying functions or sets satisfying specific combinatorial or analytic constraints~(e.g., \cite{matolcsi2010improved, vinuesa2010generalized, haugland2016minimum, gyarmati2007sums}). Progress often relies on finding a single construction that surpasses all previously known examples, thereby establishing new lower or upper bounds for the optimal value. We demonstrate that \method serves as a powerful tool for exploring the vast search space inherent in these problems, successfully tackling a diverse array of open mathematical challenges.

To assess its capabilities, we apply \method to a curated set of over 50 mathematical problems, spanning more than five different branches of mathematics, including analysis, combinatorics, number theory, and geometry, evaluated across numerous specific parameter settings (e.g., different dimensions or sizes). In 75\% of the cases \method rediscovered the best known constructions, and in 20\% of the cases it discovered a new object that is better than a previously known best construction, thereby improving the SOTA. In all these cases, the initial starting point was a simple or a random construction. These results underscore \method's broad potential as a versatile tool for mathematical research.

\begin{figure}[t]
    \centering
    \includegraphics[width=1.0\textwidth]{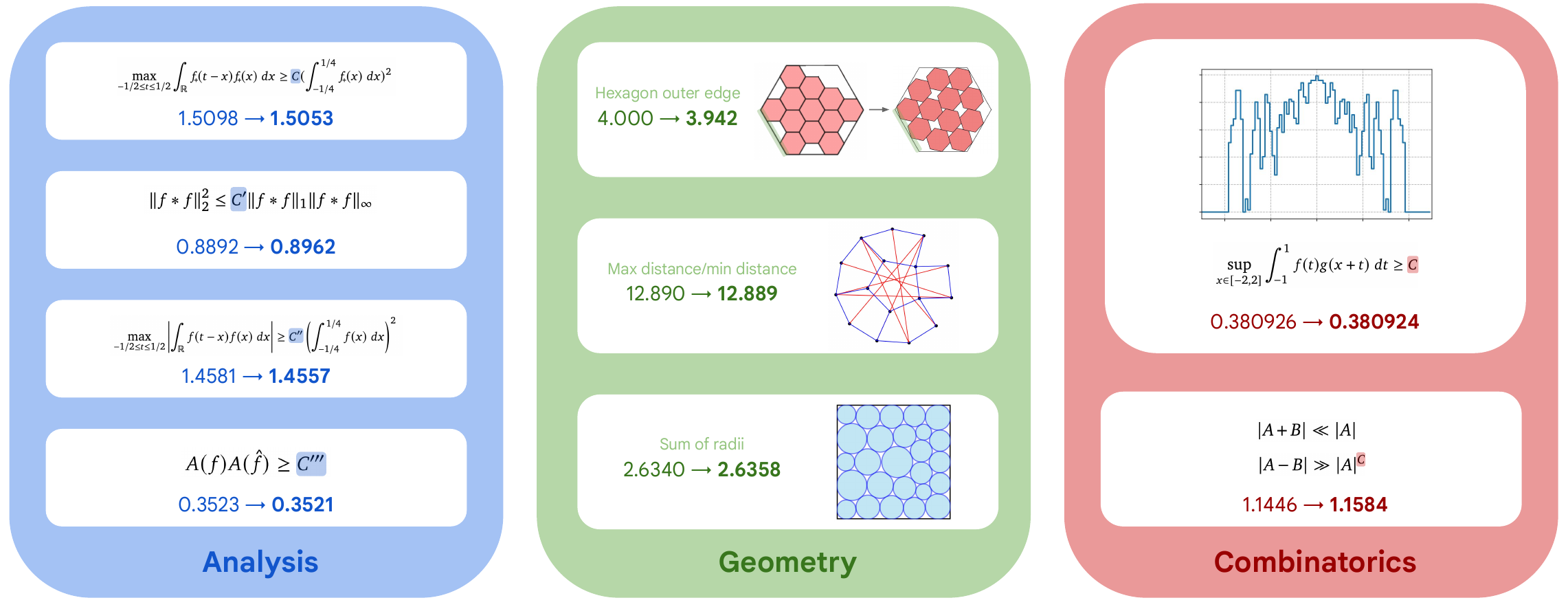}
    \caption{Examples of SOTA-breaking mathematical constructions discovered with \method. The versatility of \method allows us to tackle problems in analysis (autocorrelation and uncertainty inequalities), geometry (packing and minimum/maximum distance problems) and combinatorics (Erd\H{o}s's minimum overlap problem and sums and differences of finite sets).}
    \label{fig:math_sota_examples}
\end{figure}

\newpage % to avoid splitting the paragraph into 3 pages.
A significant advantage of the \method configuration used here is its versatility and speed of application. The core methodology, focused on evolving heuristic search programs (detailed below), can be rapidly deployed across a diverse range of mathematical construction problems and conjectures, often requiring less initial problem-specific expert tailoring compared to traditional bespoke approaches. While deep mathematical insight naturally aids in problem formulation and search space definition, \method often demonstrates a capacity to autonomously discover effective search patterns and attack strategies by identifying subtle structures within the problem landscape. This allows for efficient, large-scale exploration across many different problems.

The key methodological innovation enabling these discoveries is \method's ability to evolve \textit{heuristic search algorithms} rather than directly evolving the constructions themselves. For many problems, particularly those with fast objective function evaluations---which are common in mathematics---we employed an iterative refinement strategy. Each generation of \method was tasked with evolving a program representing a search heuristic. This program was given a fixed time budget (e.g., 1000 seconds) and was shown the best construction found by the previous best heuristic. Its goal was to leverage this starting point and the allotted time to find an even better construction. The evolutionary process thus selects for heuristics that are effective at improving already high-quality solutions. The final constructions were often the result of a sequence of different, specialized heuristics discovered by \method---early heuristics proficient at making large gains from random or simple initial states, and later heuristics adept at fine-tuning near-optimal configurations. This automated discovery of multi-stage, adaptive search strategies is challenging to replicate manually and proved crucial for surpassing the SOTA.

Below are high-level descriptions of some of the problems where \method yielded new results. Full list of problems and details are provided in Appendix~\ref{app:maths}.

\begin{itemize}
    \item \textbf{Analysis} 
    \begin{itemize}
        \item \textbf{Autocorrelation inequalities.} \method was able to improve the best known bounds on several autocorrelation inequalities.
        \item \textbf{Uncertainty principles.} \method was able to produce a refined configuration for a problem arising in Fourier analysis, by polishing an uncertainty principle construction \cite{gonccalves2017hermite} leading to a slightly better upper bound. 
    \end{itemize}

    \item \textbf{Combinatorics and number theory}
        \begin{itemize}
            \item \textbf{Erd\H{o}s's minimum overlap problem.} \method established a new upper bound for the minimum overlap problem~\cite{erdHos1955some}, slightly improving upon the previous record~\cite{haugland2016minimum}.
        \end{itemize}

    \item \textbf{Geometry and packing}
        \begin{itemize}
            \item \textbf{Kissing number problem.} In 11 dimensions, \method improved the lower bound on the kissing number, finding a configuration of 593 non-overlapping unit spheres that can simultaneously touch a central unit sphere, surpassing the previous record of 592~\citep{kissing_11}. 
            \item \textbf{Packing problems.} \method achieved several new results in packing problems, such as packing $N$ points in a shape to minimize the ratio of the maximum and minimum distance, packing various polygons in other polygons in the most efficient way, and variants of the Heilbronn problem concerning point sets avoiding small-area triangles~\cite{geometry_collection}.
        \end{itemize}
\end{itemize}

The full list of problems appears in Appendix~\ref{app:maths} and the new constructions found by \method can be found in~\ResultsColab.
More examples and details on these problems and the methods used will be provided in an upcoming paper.
Most of these discoveries are on open problems suggested to us by external mathematicians Javier Gomez Serrano and Terence Tao, who also advised on how to best formulate them as inputs to \method.
This highlights the potential for synergistic partnerships between AI-driven discovery engines like \method and human mathematical expertise.

\subsection{Optimizing Google's computing ecosystem}

In addition to the scientific applications presented in preceding sections, here we demonstrate how \method has been used to improve performance of mission-critical infrastructure and deliver real-world impact.

\subsubsection{Improving data center scheduling}

Efficiently scheduling compute jobs onto a cluster of machines is a critical optimization problem, particularly at the scale of Google's data centers, orchestrated by Borg \cite{43438}. This task involves assigning jobs to available machines based on job resource requirements and machine capacity. Inefficient assignments can result in stranded resources: when a machine can no longer accept jobs because it has run out of one kind of resource (e.g., memory) but still has other resources free (e.g., CPU). Improvements in scheduling efficiency can recover these stranded resources, allowing more jobs to be completed on the same amount of computational footprint. This recovery is essential to accommodate growing compute needs without a proportional increase in resource consumption. Furthermore, this problem is challenging since it combines typical engineering difficulties, such as debuggability and scale, on top of the classically difficult bin-packing problem.

We address this challenge by framing the online job scheduling problem as a vector bin-packing problem with two variables. In this context, machines represent bins with defined capacities for CPU and memory, and incoming jobs are items with specific resource demands. A heuristic function takes as input a pending job’s CPU and memory requirements and a potential machine’s CPU and memory availability. This function then outputs a priority score for the machine. The Borg scheduler subsequently assigns the pending job to the machine with the highest priority score as determined by the heuristic function, among other objectives. Because this heuristic only influences the ranking of machines already determined by Borg to be available and capable of running each pending job, the resulting scheduling decisions are effectively correct by construction.

\begin{figure}[t]
\centering

\begin{minipage}[c]{0.55\textwidth}
\centering
\begin{minted}[
    linenos,
    breaklines=false,]{python}
def alpha_evolve_score(required, free):
  cpu_residual = required.cpu / free.cpu
  mem_residual = required.mem / free.mem

  return -1.0 * (cpu_residual + mem_residual +
                 mem_residual / cpu_residual +
                 cpu_residual / mem_residual)
\end{minted}
\end{minipage}
\hfill
\begin{minipage}[c]{0.4\textwidth}
    \includegraphics[width=1.0\textwidth]{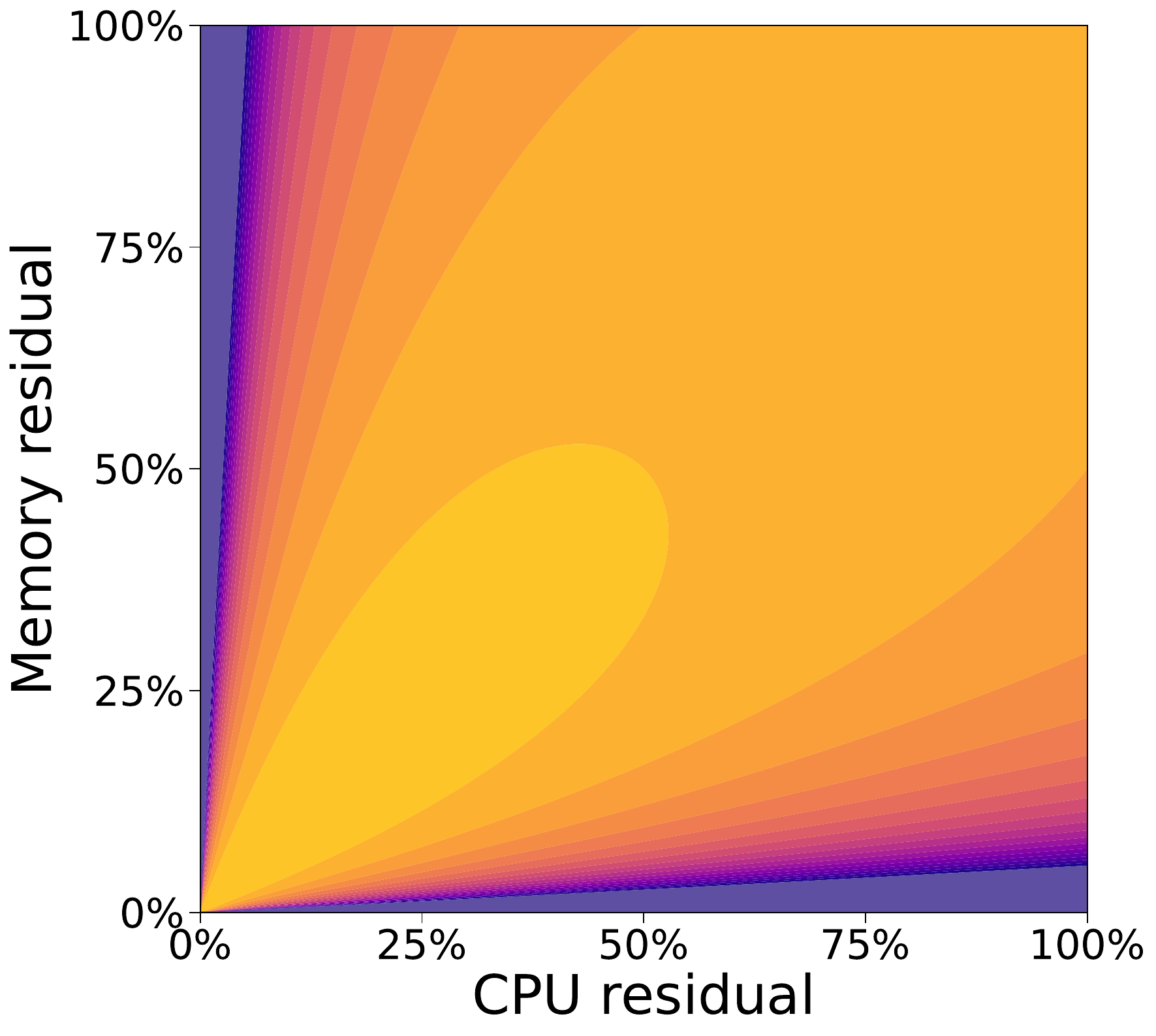}
\end{minipage}

\caption{Left: The heuristic function discovered by \method, tailored to Google’s workloads and capacity. Right: Visualization of the heuristic scoring function. Yellow regions represent high scores, while purple regions represent low scores.}
\label{fig:alphaevolve_heuristic}
\end{figure}

An early version of \method was used to discover a remarkably simple yet effective heuristic function (shown in \Cref{fig:alphaevolve_heuristic}), evolving from the existing one in production. We use a simulator of our data centers to provide feedback to \method based on historical snapshots of workloads and capacity across Google’s fleet. We measure the performance of \method's heuristic function on an unseen test dataset of recent workloads and capacity to ensure generalization. Observing that \method's heuristic function outperforms the one in production, we rolled out \method's heuristic function to the entire fleet. Post-deployment measurements across Google’s fleet confirmed the simulator results, revealing that this heuristic function continuously recovers on average 0.7\% of Google’s fleet-wide compute resources, which would otherwise be stranded. \method was chosen over a deep reinforcement learning approach because its code solution not only leads to better performance, but also offers clear advantages in interpretability, debuggability, predictability, and ease of deployment—essential qualities for a mission-critical system.

\subsubsection{Enhancing Gemini kernel engineering}

\begin{figure}[t]
    \centering
    \includegraphics[width=0.5\textwidth]{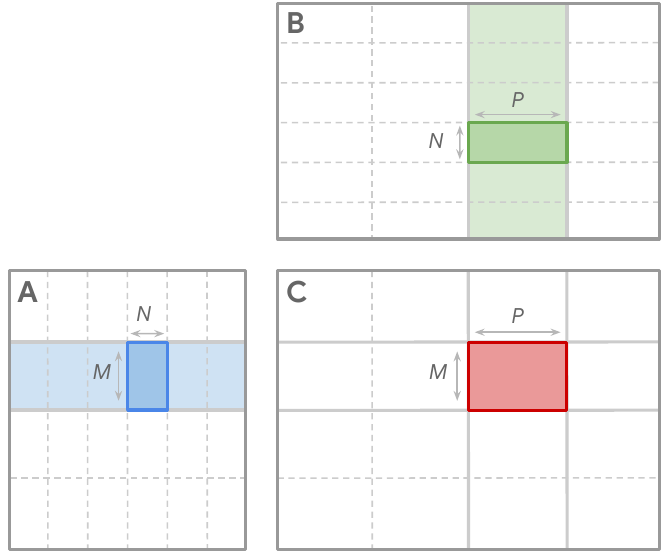}
    \caption{Visualization of the tiling heuristic problem for a matrix  product $AB = C$. Creating a heuristic that automatically chooses the right tile size ($M$, $N$, $P$) for all input shapes is difficult because one has to know the matrix multiplication unit’s optimal shapes and memory capacity, the memory requirements of surrounding operations, extra operations that are fused into the kernel, and low-level compiler intricacies, among other details.}
    \label{fig:tiling_heuristic}
\end{figure}

Training large models like Gemini requires substantial computational resources. Gemini is built on JAX \cite{jax2018github}, and Pallas is an extension to JAX that enables writing custom, highly specialized programs (kernels) tailored for optimal execution on hardware accelerators. Therefore, efficient Pallas kernels are crucial for optimizing Gemini’s training performance. A critical aspect of kernel optimization is tuning the tiling strategy for matrix multiplication operations (see~\Cref{fig:tiling_heuristic}). This technique involves dividing a large matrix multiplication computation into smaller subproblems to better balance computation with data movement, which is key to accelerating the overall computation. Traditionally, kernel engineers rely on either search-based autotuning or manually crafted heuristics to determine near-optimal tiling configurations for various input shapes. Search-based tuning interrupts the research workflow, necessitating retuning for every input shape change. Conversely, manually crafting effective tiling heuristics is a major engineering bottleneck due to its complexity, demanding a deep understanding of both kernel functionality and hardware intricacies. The key advantage of a performant heuristic is its ability to deliver high performance across arbitrary input shapes. Consequently, to expedite the design of performant kernels for emerging hardware and to simplify their utilization by model developers, we aim to facilitate the heuristic generation process.

We address this challenge by employing \method to optimize tiling heuristics for an important matrix multiplication kernel used to train Gemini. The objective is to minimize the kernel's actual runtime. \method iteratively explores and refines tiling heuristics for this kernel by proposing candidate code, aiming to minimize this runtime on various input shapes on real TPU accelerators. The kernel’s correctness is maintained by construction because \method is optimizing the tiling strategy for this kernel rather than altering its underlying mathematical operation. To build the training and evaluation datasets for \method, we automatically collect realistic kernel input shapes from kernel users. Half of these input shapes form the training set, providing the optimization targets during the evolutionary process. The remaining input shapes form the evaluation set, used to test the general applicability of the resulting heuristic.

This automated approach enables \method to discover a heuristic that yields an average 23\% kernel speedup across all kernels over the existing expert-designed heuristic, and a corresponding 1\% reduction in Gemini’s overall training time. In addition, the use of \method significantly reduced the kernel optimization time, from several months of dedicated engineering effort to just days of automated experimentation. This acceleration speeds up the deployment of optimized kernels, allowing kernel engineers to dedicate their expertise to more strategic, higher-level optimization problems. Furthermore, \method offers a path towards automating the manual tuning process and improving the ergonomics of Gemini kernel usage. The tiling heuristic discovered by \method has been deployed in production, directly enhancing Gemini's training efficiency and the Gemini team’s research and engineering velocity. This deployment also marks a novel instance where Gemini, through the capabilities of \method, optimizes its own training process.

\subsubsection{Assisting in hardware circuit design}

Specialized hardware, such as Google's Tensor Processing Units (TPUs), is crucial for achieving the resource efficiency required to run modern AI systems at scale. However, designing new computer chips is a complex and time-consuming process, often spanning years. Register-Transfer Level (RTL) optimization, a critical step in this process, involves manually rewriting hardware descriptions to improve metrics like power, performance, and area, demanding months of iteration by highly skilled engineers.

In this work, \method was challenged to optimize an already highly optimized Verilog implementation of a key TPU arithmetic circuit within the matrix multiplication unit. The optimization objectives were to reduce both area and power consumption while preserving the component's core functionality. Crucially, the final proposal must pass robust verification methods to confirm that the modified circuit maintains functional correctness. \method was able to find a simple code rewrite that removed unnecessary bits, a change validated by TPU designers for correctness. While this specific improvement was also independently caught by downstream synthesis tools, \method's contribution at the RTL stage demonstrates its capability to refine source RTL and provide optimizations early in the design flow.

Integrated into an upcoming TPU, this improvement represents Gemini’s first direct contribution to TPU arithmetic circuits, achieved via \method, paving the way for future contributions. A key advantage of \method is that it communicates the suggested changes directly in Verilog, the standard language used by hardware engineers, fostering trust and simplifying adoption. This early exploration demonstrates a novel approach where LLM-powered code evolution assists in hardware design, potentially reducing time to market.

\subsubsection{Directly optimizing compiler-generated code}

The transformer architecture~\cite{vaswani2017attention} is used in the majority of modern neural networks, ranging from LLMs to AlphaFold~\cite{alphafold3}. 
The core computation of transformers is the attention mechanism~\cite{bahdanau2014neural}, which is most commonly implemented using FlashAttention~\cite{dao2022flashattention}.
In our stack, FlashAttention is implemented as an accelerator kernel in Pallas, wrapped by higher-level code in JAX that handles input preparation and output postprocessing.
The machine learning compiler (XLA~\cite{xla}) then translates this implementation into a sequence of intermediate representations (IRs), each adding more detail for execution on particular hardware. At these stages, improved decisions on memory access orchestration or computation scheduling can significantly reduce runtime on specific hardware.

We challenged \method to directly optimize the XLA-generated IRs encapsulating the FlashAttention kernel along with pre- and postprocessing code. We optimized a configuration corresponding to a highly impactful transformer model used for inference at scale on GPUs, with the goal of minimizing the module's overall execution time. This was a particularly challenging task, because (1) the IR is designed for debugging purposes rather than for direct editing by developers, and (2) it is compiler-generated and already highly optimized. Each modification proposed by \method was checked against the reference (unmodified) code on randomized inputs in order to ensure numerical correctness throughout optimization. The final version of the code was rigorously confirmed by human experts to be correct for all possible inputs.

\method was able to provide meaningful optimizations for both levels of abstraction exposed by the IR. Firstly, the FlashAttention kernel for the configuration of interest was sped up by 32\%. Secondly, \method found improvements in pre- and postprocessing of kernel inputs and outputs, resulting in a 15\% speed up in this part. These results demonstrate the ability of \method to optimize compiler-generated code, offering the potential of incorporating discovered optimizations into existing compilers for specific use cases, or, in the longer term, incorporating \method into the compiler workflow itself.

\section{Ablations}
\label{sec:ablations_rewrite}

We carried out ablations on two tasks: finding tensor decompositions for faster matrix multiplication (\Cref{subsec:matmul}) and computing lower bounds on kissing numbers (\Cref{subsec:math}), aiming to understand the efficacy of the following components of \method.

\begin{itemize}
    \item {\bf Evolutionary approach.} \method utilizes an evolutionary approach, where previously generated programs are stored in a database and used to obtain better programs in subsequent iterations. To analyze the importance of evolution, we consider an alternative approach, which repeatedly feeds the same initial program to the language model. We refer to this approach as ``No evolution''.
    \item {\bf Context in prompts.} \method uses powerful language models with large context windows, whose output can be improved significantly by providing problem-specific context in the prompt. To test the importance of context, we consider an alternative approach where no explicit context is added to the prompt. We refer to this approach as ``No context in the prompt''.
    \item {\bf Meta prompts.} \method also uses meta prompts in order to improve the prompts that are provided to the language model. This allows it to potentially surpass the performance one can obtain using a human prompter. To test the efficacy of meta prompting, we disable it for the task of tensor decomposition. We refer to this approach as ``No meta prompt evolution''.
    \item {\bf Full-file evolution.} Unlike previous approaches such as FunSearch, \method can evolve an entire codebase instead of focusing on a single function. To test the importance of full-file evolution, we consider an alternative in the context of tensor decomposition where only the loss function is evolved. We refer to this approach as ``No full-file evolution''.
    \item {\bf Powerful language models.} \method relies on a mixture of small and large language models in order to obtain highly diverse samples. To understand the importance of this component, we consider an alternative where only a single small base model is used. We refer to this approach as ``Small base LLM only''.
\end{itemize}

\begin{figure}
\centering
\begin{subfigure}[c]{0.48\textwidth}
    \centering
    \includegraphics[width=1.0\textwidth]{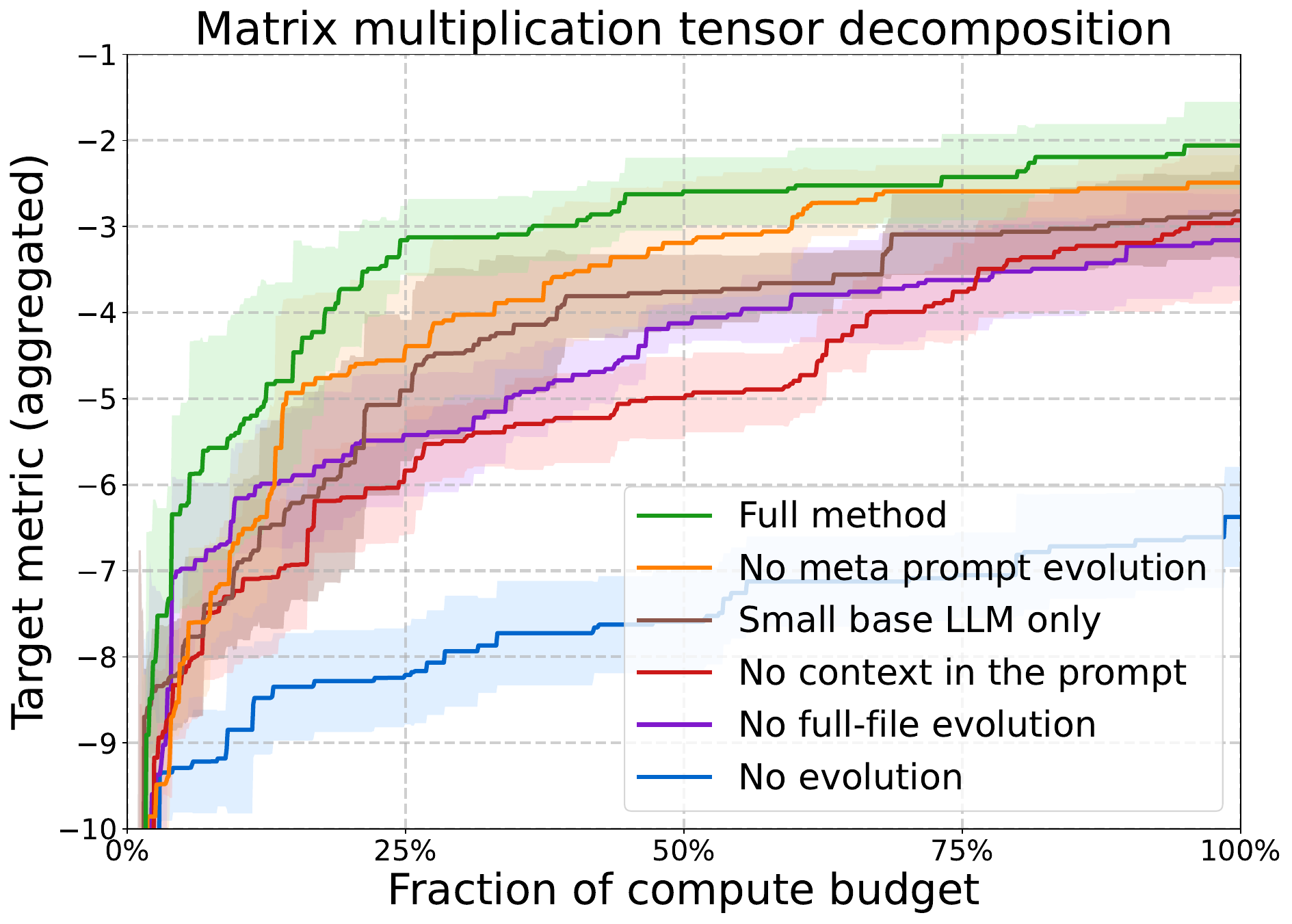}
\end{subfigure}%
\hfill
\begin{subfigure}[c]{0.48\textwidth}
    \centering
    \includegraphics[width=1.0\textwidth]{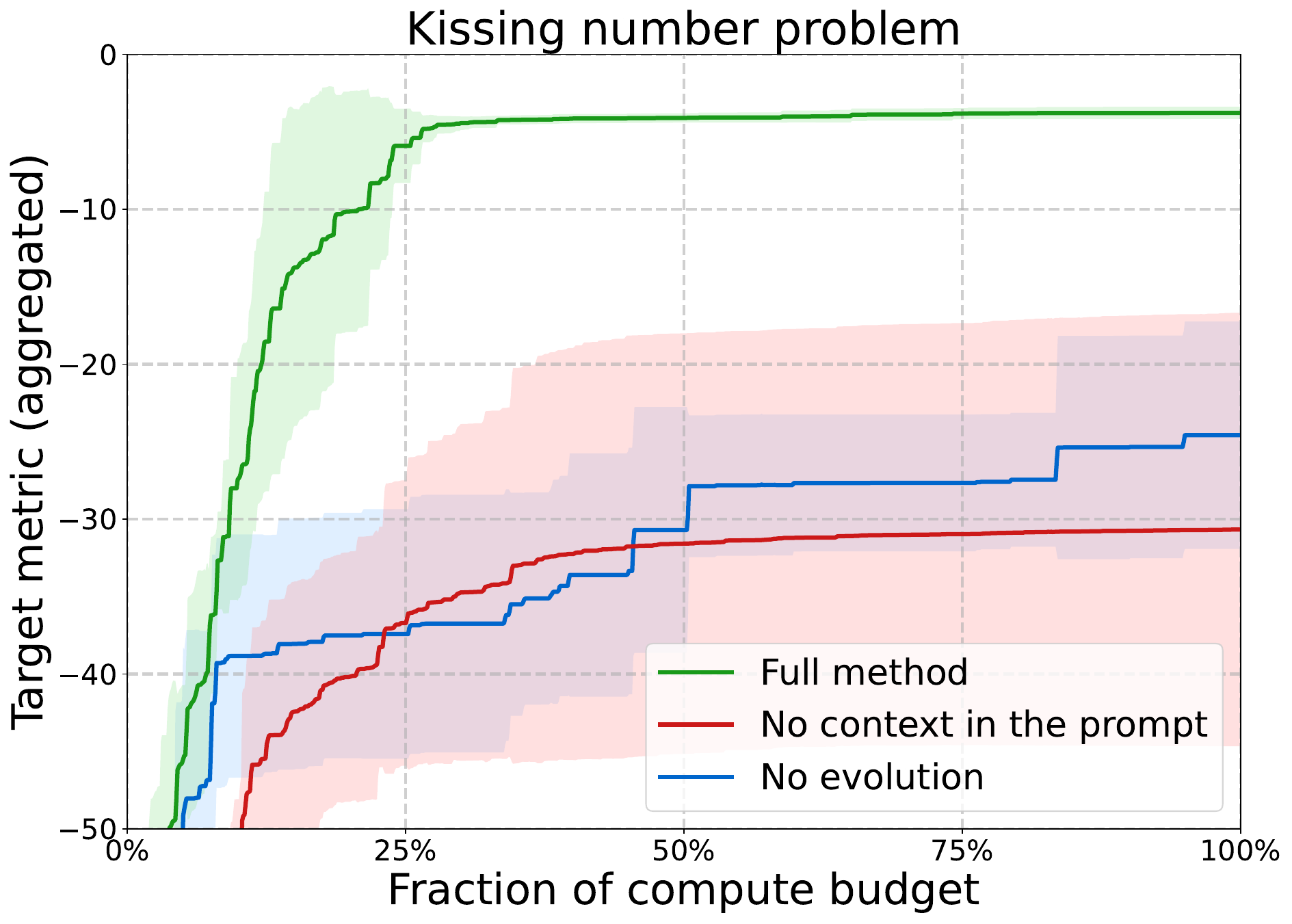}
\end{subfigure}%
\caption{Left: Ablations of \method on the problem of finding low-rank tensor decomposition for faster matrix multiplication. Right: Ablations of \method on the problem of finding sphere packings for improving kissing numbers. Each curve shows the performance of an individual setting with increasing compute budget, averaged over all considered targets (higher values on the target metric are better). The shades indicate intra-target standard deviation, averaged over three independent runs of \method, initialized with different random seeds.%
\label{fig:ablations_rewrite}}
\end{figure}

\Cref{fig:ablations_rewrite} shows the results of the all-inclusive \method approach as well as the various alternatives listed above. As can be seen, each of the components is responsible for a significant improvement in the results.

\section{Related work}
\label{sec:related_work}

\paragraph{Evolutionary methods.}
\method extends a long tradition of research on \emph{evolutionary} or \emph{genetic programming} \citep{langdon2013foundations}, where one repeatedly uses a set of mutation and crossover operators to evolve a pool of programs \citep{koza1994genetic,banzhaf1998genetic}.
In particular, classical evolutionary techniques have succeeded in symbolic regression applications \citep{schmidt2009distilling,ma2022evolving}, automated scientific \citep{cranmer2023interpretable} or algorithmic \citep{chen2023symbolic} discovery, and scheduling \citep{zhang2021genetic} problems.
However, a challenge with these methods is the use of handwritten evolution operators, which can be hard to design and may fail to capture important properties of the domain.
In contrast, \method uses LLMs to automate the construction of these operators---it leverages the LLM's world knowledge to mutate programs without the need to pre-define a set of allowed mutation operations. 

\method was preceded by a body of recent efforts that combine LLMs and evolution; specifically, it extends the FunSearch system, introduced by \citet{paredes2023mathematical} as an approach to mathematical discovery.
FunSearch was subsequently used in downstream tasks such as learning acquisition functions for Bayesian optimization \citep{aglietti2025funbo}, discovering cognitive models \citep{castro2025discovering}, computing distances between graphs \citep{verma2025grail}, or combinatorial competitive programming \citep{velickovic2024amplifying}. 
\method goes beyond FunSearch and its recent reimplementation \citep{ellenberg2025generative} in three key ways.
First, while FunSearch only allowed the evolution of a single Python function, \method allows evolution over entire codebases written in a wide range of programming languages.
Second, FunSearch optimized a single objective function, while \method provides the ability to perform multiobjective optimization.
Third, the LLMs in FunSearch were relatively small and solely trained on code.
By contrast, \method uses frontier LLMs and rich forms of natural-language context and feedback.
As has been demonstrated in this paper, these extensions allow \method to address important challenging problems that were not amenable to FunSearch.

Other efforts in this category include the approach by \citet{lehman2023evolution}, which uses an LLM-guided evolution process to discover programmatic policies for a set of simulated robots; or the approach by \citet{hemberg2024evolving} for code synthesis.
Similar approaches have found use in several scientific and mathematical tasks, including symbolic regression \citep{shojaee2025llmsr,grayeli2024symbolic}, discovering heuristics for combinatorial optimization \citep{liu2024evolution,ye2024reevo,yao2025multiobjective}, and synthesizing molecular structures \citep{wang2025efficient}.
LLM-guided evolution has also been used to improve AI systems by enhancing LLM prompts \citep{fernando2023promptbreeder} and searching over neural architectures \citep{chen2023evoprompting,morris2024llm}.
\method differs from these approaches in its scale, flexibility, and general applicability to a broad range of domains.

Some recent efforts have augmented the basic paradigm of LLM-guided evolution with complementary ideas.
For example, \citet{surina2025algorithm} complement the evolution process by continuously finetuning the LLM through reinforcement learning.
\citet{grayeli2024symbolic} enhance the evolution process with an LLM-directed concept learning step that summarizes high-performing programs in the pool into natural language.
More investigation is required to understand the benefits of these ideas at the scale at which \method operates. 

Evolutionary methods have also found use in the recent AI Co-Scientist work~\citep{gottweis2025towards}, which seeks to automate scientific discovery using distinct agents for tasks like hypothesis discovery, ranking of hypotheses, and literature review.
While AI Co-Scientist represents scientific hypotheses and their evaluation criteria in \emph{natural language}, \method focuses on evolving \emph{code}, and directs evolution using programmatic evaluation functions.
This choice enables us to substantially sidestep LLM hallucinations, which allows \method to carry on the evolution process for a large number of time steps.
Nevertheless, it is possible in principle to combine the two approaches, leading to a method that allows a flexible combination of natural-language and programmatic idioms. 

\paragraph{Superoptimization and algorithm discovery.}
\method can be viewed as a method for \emph{code superoptimization} in that it iteratively improves an initial program using execution feedback. 
The idea of code superoptimization goes back to the 1980s \citep{Massalin87}; pre-LLM approaches to the problem included systematic enumeration \citep{Massalin87}, genetic search \citep{cooper2002adaptive}, Monte Carlo sampling \citep{schkufza2013stochastic}, and deep reinforcement learning \citep{mankowitz2023faster}.
Additionally, in limited settings that focus on a single problem such as matrix multiplication, there have been systems such as AlphaTensor that were also able to discover provably correct algorithms \citep{fawzi2022discovering}.

More recently, a body of LLM-based approaches to superoptimization and algorithm discovery have emerged.
This literature builds on the success of LLMs in coding tasks, perhaps best illustrated by their success in (simulated) programming competitions as in the case of AlphaCode \citep{li2022competitionlevel}.
For instance, LLM agents have been used to optimize certain operations in GPU kernels, such as the attention operation \citep{chen2025automating} or more general user-specified operations \citep{lange2025aicuda}.
There is also work on using LLMs to discover novel evolutionary algorithms \citep{lange2024large}, train language models \citep{lehman2024evolution}, and optimize warehouse-scale computers \citep{lin2025eco}.
Other recent work \citep{wu2023autogen} has also proposed the use of multiple LLM agents that converse with each other to accomplish mathematical and coding tasks. 

While previous work on using LLMs for algorithm discovery provided promising results, \method's approach to leverage it for evolutionary algorithms allows us to address significantly more challenging problems, as demonstrated in \Cref{sec:results}.

\paragraph{AI for scientific and mathematical discovery.}
Over the last decade, AI systems have been applied to a wide range of scientific disciplines and tasks, from protein structure prediction \citep{jumper2021highly} to quantum physics \citep{bausch2024learning,ruiz2025quantum} to climate sciences \citep{lam2023learning}.
In particular, there are numerous recent LLM-based methods that target scientific problems in multiple disciplines, such as materials science \citep{miret2024llms,zhang2024honeycomb,jia2024llmatdesign,song2025llmfeynman}, chemistry \citep{caldasramos2025review,luo2025leveraging}, bioinformatics \citep{sarumi2024large,madani2023large}, geoscience \citep{pantiukhin2025accelerating}, and quantum physics \citep{frohnert2025discovering,pan2025quantum} (for surveys on the topic, see \citep{gridach2025agentic,luo2025llm4sr,ren2025towards}).

Many of these methods use LLMs to automate several distinct stages of the scientific discovery process \citep{wang2024scimon,xia2025nature,li2025large,gu2024interesting,yang2024large}, e.g., for generating and ranking hypotheses and ideas \citep{guo2024embracing,si2025canllms}.
Of these methods, especially related to \method are the methods that use LLM-guided tree search-based algorithms \citep{bran2025chemical} or LLM-guided evolutionary algorithms \cite{yang2025moosechem,zhou2024hypothesis,gottweis2025towards}.
Other works use LLMs to optimize experimental planning and design \citep{huang2024crisprgpt,bran2024augmenting,boiko2023autonomous,naumov2025dora} or experiment execution and workflow \citep{wang2025efficient,rives2021biological,lin2023evolutionary,ye2023drugassist,ferruz2022controllable}.
Finally, there are also works focusing on the data analysis stage \citep{rasheed2024canlarge}.
\method differs from most of these methods in its use of programmatic hypothesis representations and evaluation metrics.

AI systems have also contributed to advances in pure mathematics \citep{davies2021advancing}.
In this context, the FunSearch approach \citep{paredes2023mathematical,ellenberg2025generative} established LLM-guided evolution as a powerful tool for discovering witnesses for, and counterexamples to, mathematical statements---a problem that is complementary to that of finding formal and informal proofs of mathematical statements~\citep{trinh2024solving,hubert2024ai,yang2024formal,yang2023leandojo,thakur2024context,collins2024evaluating}.

\section{Discussion}
\label{sec:discussion}

\method demonstrates the surprising power of combining state-of-the-art LLMs with automated evaluation metrics within an evolutionary framework, which can lead to new discoveries on decades-old mathematical problems as well as practical improvements to highly optimized compute stacks.

Interestingly, \method often allows approaching the same problem in different ways: searching for the solution directly, finding a function that constructs it from scratch, or evolving a search algorithm to find it.
Applying \method in different ways  comes with different biases (for example, finding constructive functions may favor discovering highly symmetric objects~\cite{paredes2023mathematical}) and thus can suit different problems.

\method can also be seen as a test-time compute agent that, through its evolutionary procedure, significantly enhances the capability of the base LLM (compared to, e.g., repeated sampling).
On one hand, this can be seen as a compelling demonstration of how machine feedback is able to sustain test-time compute scaling up to regimes where new scientific discoveries and highly valuable practical optimizations are made.
On the other hand, a natural next step will be to consider distilling the \method-augmented performance of the base LLMs into the next generation of the base models.
This can have intrinsic value and also, likely, uplift the next version of \method.

Beyond distillation, it is also intriguing that \method can make practical discoveries that increase the efficiency of its own infrastructure and of (future versions of) its base LLMs.
Currently, the gains are moderate and the feedback loops for improving the next version of \method are on the order of months.
However, with these improvements we envision that the value of setting up more environments (problems) with robust evaluation functions will become more widely recognized, which in turn will result in more high-value practical discoveries going forward.

The main limitation of \method is that it handles problems for which it is possible to devise an automated evaluator.
While this is true of many problems in the mathematical and computational sciences, there are domains such as the natural sciences where only some experiments can be simulated or automated.
While \method does allow for LLM-provided evaluation of ideas, this is not a setting we have optimized for.
However, concurrent work shows this is possible~\cite{gottweis2025towards}, and a natural step would be to link the two settings, with LLMs providing feedback on high-level ideas before transitioning to an implementation stage, for which machine feedback is available through code execution.

\section*{Acknowledgements}
\label{sec:acknowledgements}

We thank  Michael Figurnov for reviewing this white paper; Alhussein Fawzi, Bernardino Romera-Paredes, and Ankit Anand for early explorations and insightful discussions; Stig Petersen and Demis Hassabis for support and advice;  JD Velasquez for helpful advice on managing the practical applications; and all early users and collaborators of \method for their diverse use cases and insightful feedback, which shaped it into a more robust and versatile tool for a wide range of applications. We gratefully acknowledge the invaluable contributions of these individuals towards the applications highlighted in this white paper:

Terence Tao, Javier Gomez Serrano, and Jordan Ellenberg for suggesting specific open mathematical problems and advising on how to best formulate them for \method; Bogdan Georgiev, Ray Jiang, and Johannes Bausch for their contributions to applying \method to such problems.

Mohammadamin Barekatain, Patrick Heisel, Chase Hensel, Robert O'Callahan, and Pengming Wang for co-leading the application to data center scheduling; Federico Piccinini, Sultan Kenjeyev, and Andrea Michi for making significant contributions; Kieran Milan, Daniel Mankowitz, Cosmin Paduraru, Calin Cascaval, Tammo Spalink, and Natasha Antropova for providing helpful advice; Aaron Gentleman, Gaurav Dhiman, Parthasarathy Ranganatha, and Amin Vahdat for reviewing this work.

Yanislav Donchev for leading the application to Gemini kernel engineering; Richard Tanburn for making significant contributions; Justin Chiu and Julian Walker for providing helpful advice; Jean-Baptiste Alayrac, Dmitry Lepikhin, Sebastian Borgeaud, Koray Kavukcuoglu and Jeff Dean for reviewing this work.

Timur Sitdikov for leading the application to TPU circuit design; Georges Rotival for providing the circuit evaluation infrastructure; Kirk Sanders, Srikanth Dwarakanath, Indranil Chakraborty, Christopher Clark for verifying and validating the results in the TPU design; Vinod Nair, Sergio Guadarrama, Dimitrios Vytiniotis, and Daniel Belov for their helpful advice; Kerry Takenaka, Jeff Dean, Sridhar Lakshmanamurthy, Parthasarathy Ranganathan, and Amin Vahdat for reviewing this work.

Benjamin Chetioui, Sergei Lebedev, Alexander Belyaev, Henning Becker, Oleg Shyshkov, and Aliia Khasanova for their help with XLA modifications as well as their helpful advice; Giorgio Arena, Marco Cornero, and Sebastian Bodenstein for reviewing this work.

\section*{Author information}
\label{sec:authors}

These authors contributed equally: Alexander Novikov, Ngân Vũ, Marvin Eisenberger, Emilien Dupont, Po-Sen Huang, Adam Zsolt Wagner, Sergey Shirobokov, Borislav Kozlovskii, and Matej Balog.

\paragraph{Contributions.}
A.N. and M.B. designed and implemented the initial version of \method.
\;
M.B., A.N., N.V. and P.K. developed project vision and scoped problems.
\;
N.V. and P.-S.H. oversaw the practical applications.
\;
E.D. and M.E. implemented the first benchmark problem used for iterating on \method, with input from F.J.R.R. and M.B.
\;
A.N. and M.E. developed the final version of \method, with contributions from S.S., P.-S.H., and input from M.B., E.D., A.Z.W. and N.V.
\;
A.N., S.S., P.-S.H. and M.E. maintained the infrastructure underlying \method.
\;
M.E. and E.D. used \method to discover new algorithms for matrix multiplication, with input from F.J.R.R.
\;
A.Z.W. worked on the applications to open mathematical problems, with help from A.M., M.E., and A.N.
\;
A.N. contributed to the Borg scheduling application.
\;
P.-S.H. and N.V. worked on the application to Gemini kernel engineering.
\;
P.-S.H. and A.N. contributed to the TPU circuit design application.
\;
B.K. and S.S. worked on applying \method to directly optimize compiler-generated code.
\;
M.E. performed the ablation experiments.
\;
M.B., A.N., M.E., S.S. and P.-S. H. performed the majority of code reviews.
\;
M.B., E.D., S.C., N.V., A.Z.W., F.J.R.R., M.E., A.N., B.K., S.S., A.M., and M.P.K. wrote the paper, with input from A.S.,  P.-S.H and P.K.
\;
N.V., E.D., M.E., S.C., A.N., and A.Z.W. created the figures.
\;
F.J.R.R., A.M., and A.Z.W. assembled the accompanying Google Colab.
\;
S.N., A.D. and P.K. advised and enabled multiple strands of this work.
\;
M.B., A.N., N.V. and G.H. coordinated the team. 
\;
P.K. supervised and coordinated the research program. 

\paragraph{Corresponding authors.}
Matej Balog, Alexander Novikov and Pushmeet Kohli.

\bibliographystyle{abbrvnat}
\bibliography{literature}

\clearpage
\appendix

\section{Faster matrix multiplication: Full results}

\paragraph{Full table of results.} We provide the best ranks obtained by \method in~\Cref{tab:relaxed-opt-results-appendix}. Overall, we considered 54 matrix multiplication sizes in our experiments. These were chosen roughly representing sizes $\langle m,n,p \rangle$ where $2\leq m,n\leq 5$, with some reasonable cutoff for $p$. Due to symmetries of the underlying matrix multiplication tensor, there exist equivalent algorithms for any permutations of the three axes, hence we focus on sorted sizes $m\leq n\leq p$.

In all but two considered sizes, \method discovered programs which either match or surpass the best known rank. Anecdotally, we encountered some difficulty when increasing the problem size: when we run the discovered programs on sizes beyond $\langle 5,5,5 \rangle$ on 1000 random seeds on evaluators with a single GPU accelerator, we often run out of memory. Hence, extending our setup to larger matrix sizes requires further optimization.

\begin{table}[h]
\rowcolors{2}{white}{gray!30}
\begin{center}
    \scalebox{0.7}{
    \begin{tabular}{ccc}
    $\langle m, n, p \rangle$ & \makecell{best known \\ {[reference]}}  & \method \\ \midrule
    $\langle 2, 2, 2 \rangle$ & 7 \citep{strassen1969gaussian}   & 7   \\ 
    $\langle 2, 2, 3 \rangle$ & 11~\citep{smirnov2013bilinear}& 11 \\ 
    $\langle 2, 2, 4 \rangle$ & 14~\citep{smirnov2013bilinear} & 14 \\ 
    $\langle 2, 2, 5 \rangle$ & 18~\citep{smirnov2013bilinear} & 18 \\ 
    $\langle 2, 2, 6 \rangle$ & 21~\citep{smirnov2013bilinear} & 21 \\ 
    $\langle 2, 2, 7 \rangle$ & 25~\citep{smirnov2013bilinear} & 25 \\ 
    $\langle 2, 2, 8 \rangle$ & 28~\citep{smirnov2013bilinear} & 28 \\ 
    $\langle 2, 2, 9 \rangle$ & 32~\citep{smirnov2013bilinear} & 32 \\ 
    $\langle 2, 2, 10 \rangle$ & 35~\citep{smirnov2013bilinear} & 35 \\ 
    $\langle 2, 2, 11 \rangle$ & 39~\citep{smirnov2013bilinear} & 39 \\ 
    $\langle 2, 2, 12 \rangle$ & 42~\citep{smirnov2013bilinear} & 42 \\ 
    $\langle 2, 2, 13 \rangle$ & 46~\citep{smirnov2013bilinear} & 46 \\ 
    $\langle 2, 2, 14 \rangle$ & 49~\citep{smirnov2013bilinear} & 49 \\ 
    $\langle 2, 2, 15 \rangle$ & 53~\citep{smirnov2013bilinear} & 53 \\ 
    $\langle 2, 2, 16 \rangle$ & 56~\citep{smirnov2013bilinear} & 56 \\ 
    $\langle 2, 3, 3 \rangle$ & 15~\citep{smirnov2013bilinear} & 15 \\ 
    $\langle 2, 3, 4 \rangle$ & 20~\citep{smirnov2013bilinear} & 20 \\ 
    $\langle 2, 3, 5 \rangle$ & 25~\citep{smirnov2013bilinear} & 25 \\ 
    \end{tabular}
    }
    \scalebox{0.7}{
    \begin{tabular}{ccc}
    $\langle m, n, p \rangle$ & \makecell{best known \\ {[reference]}}  & \method \\ \midrule
    $\langle 2, 3, 6 \rangle$ & 30~\citep{smirnov2013bilinear} & 30 \\ 
    $\langle 2, 3, 7 \rangle$ & 35~\citep{smirnov2013bilinear} & 35 \\ 
    $\langle 2, 3, 8 \rangle$ & 40~\citep{smirnov2013bilinear} & 40 \\ 
    $\langle 2, 3, 9 \rangle$ & 45~\citep{smirnov2013bilinear} & 45 \\ 
    $\langle 2, 3, 10 \rangle$ & 50~\citep{smirnov2013bilinear} & 50 \\ 
    $\langle 2, 4, 4 \rangle$ & 26~\citep{smirnov2013bilinear} & 26 \\ 
    $\langle 2, 4, 5 \rangle$ & 33~\citep{hopcroft} & \textcolor{darkgreen}{\textbf{32}} \\
    $\langle 2, 4, 6 \rangle$ & 39~\citep{smirnov2013bilinear} & 39 \\ 
    $\langle 2, 4, 7 \rangle$ & 46~\citep{smirnov2013bilinear} & \textcolor{darkgreen}{\textbf{45}} \\
    $\langle 2, 4, 8 \rangle$ & 52~\citep{smirnov2013bilinear} & \textcolor{darkgreen}{\textbf{51}} \\
    $\langle 2, 5, 5 \rangle$ & 40~\citep{smirnov2013bilinear} & 40 \\ 
    $\langle 2, 5, 6 \rangle$ & 48~\citep{smirnov2013bilinear} & \textcolor{darkgreen}{\textbf{47}} \\ 
    $\langle 3, 3, 3 \rangle$ & 23~\citep{laderman}   & 23   \\
    $\langle 3, 3, 4 \rangle$ & 29~\citep{smirnov2013bilinear} & 29 \\ 
    $\langle 3, 3, 5 \rangle$ & 36~\citep{smirnov2013bilinear} & 36 \\ 
    $\langle 3, 3, 6 \rangle$ & 40~\citep{smirnov2013bilinear} & 40 \\ 
    $\langle 3, 3, 7 \rangle$ & 49~\citep{smirnov2013bilinear} & 49 \\ 
    $\langle 3, 3, 8 \rangle$ & 55~\citep{smirnov2013bilinear} & 55 \\ 
    \end{tabular}
    }
    \scalebox{0.7}{
    \begin{tabular}{ccc}
    $\langle m, n, p \rangle$ & \makecell{best known \\ {[reference]}}  & \method \\ \midrule
    $\langle 3, 4, 4 \rangle$ & 38~\citep{smirnov2013bilinear} & 38 \\ 
    $\langle 3, 4, 5 \rangle$ & 47~\citep{fawzi2022discovering} & 47 \\ 
    $\langle 3, 4, 6 \rangle$ & 56~\citep{Kauers_2025} & \textcolor{darkgreen}{\textbf{54}} \\ 
    $\langle 3, 4, 7 \rangle$ & 66~\citep{smirnov2021} & \textcolor{darkgreen}{\textbf{63}} \\ 
    $\langle 3, 4, 8 \rangle$ & 75~\citep{smirnov2021}   & \textcolor{darkgreen}{\textbf{74}} \\ 
    $\langle 3, 5, 5 \rangle$ & 58~\citep{smirnov2021} & 58 \\ 
    $\langle 3, 5, 6 \rangle$ & 70~\citep{Kauers_2025} & \textcolor{darkgreen}{\textbf{68}} \\
    $\langle 3, 5, 7 \rangle$ & 82~\citep{smirnov2021} & \textcolor{darkgreen}{\textbf{80}} \\ 
    $\langle 4, 4, 4 \rangle$ & 49~\citep{strassen1969gaussian} & \textcolor{darkgreen}{\textbf{48}}  \\ 
    $\langle 4, 4, 5 \rangle$ & 62~\citep{kauers2023flip}   & \textcolor{darkgreen}{\textbf{61}}    \\
    $\langle 4, 4, 6 \rangle$ & 73~\citep{Kauers_2025} & 73 \\ 
    $\langle 4, 4, 7 \rangle$ & 87~\citep{smirnov2013bilinear,strassen1969gaussian} & \textcolor{darkgreen}{\textbf{85}}    \\
    $\langle 4, 4, 8 \rangle$ & 98~\citep{strassen1969gaussian} & \textcolor{darkgreen}{\textbf{96}} \\ 
    $\langle 4, 4, 9 \rangle$ & 104~\citep{smirnov2022} & \textcolor{purple}{108} \\ 
    $\langle 4, 5, 5 \rangle$ & 76~\citep{fawzi2022discovering} & 76 \\ 
    $\langle 4, 5, 6 \rangle$ & 93~\citep{Kauers_2025}   & \textcolor{darkgreen}{\textbf{90}} \\ 
    $\langle 5, 5, 5 \rangle$ & 93~\citep{flip_graphs_with_symmetry} & 93 \\ 
    $\langle 6, 6, 6 \rangle$ & 153~\citep{flip_graphs_with_symmetry} & \textcolor{purple}{156} \\ 
    \end{tabular}
    }
\caption{Full version of \Cref{tab:relaxed-opt-results}, showing the best ranks obtained by \method for tensor decomposition for all considered parameters. Of the 54 targets, \method matches the state of the art in 38 cases, surpasses it in 14 cases (green), and falls behind in 2 cases (red). In all cases, \method provides exact algorithms, using integer or half-integer entries in the decomposition. For $\langle 3, 4, 7\rangle$, $\langle 4, 4, 4\rangle$, and $\langle 4, 4, 8\rangle$, the algorithms discovered by \method use complex-valued multiplications which can be used for exact multiplication of complex or real-valued matrices. The decompositions shown in this table can be found in \ResultsColab.}
    \label{tab:relaxed-opt-results-appendix}
\end{center}
\end{table}

\emph{Note}: Concurrent work~\cite{kauers2025consequences} has also found a rank-$90$ algorithm for $\langle 4, 5, 6 \rangle$.

\paragraph{Magnified version of \Cref{fig:relaxed-opt-diff} (left).} In \Cref{fig:relaxed-opt-diff-appendix-1,fig:relaxed-opt-diff-appendix-2,fig:relaxed-opt-diff-appendix-3}, we show a magnified version of \Cref{fig:relaxed-opt-diff} (left), which corresponds to the program that discovers a decomposition of rank 48 for the 3D tensor representing the operation of multiplying two $4\times 4$ matrices. 

\begin{figure}[p]
\renewcommand{\thefigure}{\arabic{figure}a}\vspace{-0.03\textwidth}
\begin{adjustbox}{width=0.9\textwidth}
\begin{minipage}{0.9\textwidth}
\begin{lstlisting}[style=pydiff, backgroundcolor=\color{backcolour}]
@@ -45,9 +45,14 @@
   # EVOLVE-BLOCK-START
   def _get_optimizer(self) -> optax.GradientTransformation:
     """Returns optimizer."""
-    return optax.adam(self.hypers.learning_rate)
+    return optax.adamw(
+        self.hypers.learning_rate, weight_decay=self.hypers.weight_decay
+    )
 
   def _get_init_fn(self) -> jax.nn.initializers.Initializer:
     """Returns initializer function."""
-    return initializers.normal(0.0, self.hypers.init_scale, jnp.complex64)
+    # Initialize with a smaller scale to encourage finding low-rank solutions.
+    # Increase scale slightly for better exploration.
+    scale = self.hypers.init_scale
+    return initializers.normal(0 + 1j * 0, scale * 0.2, jnp.complex64)

@@ -80,6 +85,66 @@
     # Gradient updates.
     updates, opt_state = self.opt.update(grads, opt_state, decomposition)
     decomposition = optax.apply_updates(decomposition, updates)
+    # Add a small amount of gradient noise to help with exploration
+    rng, g_noise_rng = jax.random.split(rng)
+    decomposition = jax.tree_util.tree_map(
+        lambda x: x
+        + self.hypers.grad_noise_std * jax.random.normal(g_noise_rng, x.shape),
+        decomposition,
+    )
+
+    # Add noise to the decomposition parameters (exploration).
+    _, noise_rng = jax.random.split(rng)
+    noise_std = self._linear_schedule(
+        global_step, start=self.hypers.noise_std, end=0.0
+    )
+    decomposition = jax.tree_util.tree_map(
+        lambda x: x + noise_std * jax.random.normal(noise_rng, x.shape),
+        decomposition,
+    )
+
+    # Cyclical annealing for clipping threshold.
+    cycle_length = 2000  # Number of steps per cycle
+    cycle_progress = (
+        global_step % cycle_length
+    ) / cycle_length  # Normalized progress within the current cycle [0, 1)
+
+    # Map cycle progress to a sinusoidal curve. Ranges from 0 to 1.
+    clip_threshold_multiplier = (1 + jnp.cos(2 * jnp.pi * cycle_progress)) / 2
+
+    clip_threshold = self.hypers.clip_min + clip_threshold_multiplier * (
+        self.hypers.clip_max - self.hypers.clip_min
+    )
+
+    def soft_clip(x, threshold):
+      # Clipping the real and imaginary parts separately.
+      x_re = jnp.real(x)
+      x_im = jnp.imag(x)
+
+      x_re_clipped = jnp.where(
+          x_re > threshold, threshold + (x_re - threshold) * 0.1, x_re
+      )
+      x_re_clipped = jnp.where(
+          x_re_clipped < -threshold,
+          -threshold + (x_re_clipped + threshold) * 0.1,
+          x_re_clipped,
+      )
\end{lstlisting}
\end{minipage}
\end{adjustbox}
\caption{
Magnified version of~\Cref{fig:relaxed-opt-diff}(left), giving the program that discovers a faster algorithm to multiply $4\times4$ matrices (\emph{1/3}).}\label{fig:relaxed-opt-diff-appendix-1}
\centering
\end{figure}
\addtocounter{figure}{-1}

\begin{figure}[p]
\renewcommand{\thefigure}{\arabic{figure}b}\vspace{-0.03\textwidth}
\begin{adjustbox}{width=0.9\textwidth}
\begin{minipage}{0.9\textwidth}
\begin{lstlisting}[style=pydiff, backgroundcolor=\color{backcolour}, firstnumber=66]
+
+      x_im_clipped = jnp.where(
+          x_im > threshold, threshold + (x_im - threshold) * 0.1, x_im
+      )
+      x_im_clipped = jnp.where(
+          x_im_clipped < -threshold,
+          -threshold + (x_im_clipped + threshold) * 0.1,
+          x_im_clipped,
+      )
+
+      return x_re_clipped + 1j * x_im_clipped
+
+    decomposition = jax.tree_util.tree_map(
+        lambda x: soft_clip(x, clip_threshold), decomposition
+    )
+
     return decomposition, opt_state, loss
 
   def _loss_fn(
@@ -91,13 +156,86 @@
     """Computes (batched) loss on learned decomposition."""
     # Compute reconstruction loss.
     rec_tensor = self._decomposition_to_tensor(decomposition)  # (B, N, M, P)
+
+    # Add noise to the target tensor (robustness).
+    rng, noise_rng = jax.random.split(rng)
+    target_noise = self.hypers.target_noise_std * jax.random.normal(
+        noise_rng, self.target_tensor.shape
+    )
+    noisy_target_tensor = self.target_tensor + target_noise
+
+    # Hallucination loss (encourages exploration by randomly replacing values)
+    hallucination_prob = self.hypers.hallucination_prob
+    hallucination_scale = self.hypers.hallucination_scale
+
+    def hallucinate(x, hallucination_rng):
+      mask = jax.random.bernoulli(hallucination_rng, p=hallucination_prob)
+      noise = hallucination_scale * jax.random.normal(
+          hallucination_rng, x.shape
+      )
+      return jnp.where(mask, noise, x)
+
+    _, factor_rng = jax.random.split(rng)
+    decomposition = jax.tree_util.tree_map(
+        lambda x: hallucinate(x, jax.random.split(factor_rng)[0]),
+        decomposition,
+    )
+
     # Add a batch dimension to `target_tensor` to ensure correct broadcasting.
     # Define the loss as the L2 reconstruction error.
-    rec_loss = l2_loss_complex(self.target_tensor[None, ...], rec_tensor)
+    rec_loss = l2_loss_complex(noisy_target_tensor[None, ...], rec_tensor)
 
     # We must return a real-valued loss.
-    return jnp.real(rec_loss)
 
+    # Discretization loss (encourage entries to be multiples of 1/2 or integer).
+    def dist_to_half_ints(x):
+      x_re = jnp.real(x)
+      x_im = jnp.imag(x)
+      return jnp.minimum(
+          jnp.abs(x_re - jnp.round(x_re * 2) / 2),
+          jnp.abs(x_im - jnp.round(x_im * 2) / 2),
+      )
+
\end{lstlisting}
\end{minipage}
\end{adjustbox}
\caption{Magnified version of~\Cref{fig:relaxed-opt-diff}(left), giving the program that discovers a faster algorithm to multiply $4\times4$ matrices  (\emph{2/3}).}\label{fig:relaxed-opt-diff-appendix-2}
\centering
\end{figure}
\addtocounter{figure}{-1}

\begin{figure}[p]
\renewcommand{\thefigure}{\arabic{figure}c}\vspace{-0.03\textwidth}
\begin{adjustbox}{width=0.9\textwidth}
\begin{minipage}{0.9\textwidth}
\begin{lstlisting}[style=pydiff, backgroundcolor=\color{backcolour}, firstnumber=131]
+    def dist_to_ints(x):
+      return jnp.abs(x - jnp.round(x))
+
+    discretization_loss = 0.0
+    for factor in decomposition:
+      discretization_loss += jnp.mean(dist_to_half_ints(factor))
+      discretization_loss += jnp.mean(dist_to_ints(factor))
+
+    discretization_loss /= (
+        len(decomposition) * 2
+    )  # average across all factors and loss components
+
+    discretization_weight = self._linear_schedule(
+        global_step, start=0.0, end=self.hypers.discretization_weight
+    )
+
+    # Cosine annealing for half-integer loss.
+    cycle_length = self.config.training_steps // 4  # Number of steps per cycle
+    cycle_progress = (
+        global_step % cycle_length
+    ) / cycle_length  # Normalized progress within the current cycle [0, 1)
+    half_int_multiplier = (1 + jnp.cos(jnp.pi * cycle_progress)) / 2
+    half_int_multiplier = (
+        1 - self.hypers.half_int_start
+    ) * half_int_multiplier + self.hypers.half_int_start
+
+    total_loss = (
+        rec_loss
+        + discretization_weight * discretization_loss * half_int_multiplier
+    )
+
+    # Add penalty for large values (stability).
+    large_value_penalty = 0.0
+    for factor in decomposition:
+      large_value_penalty += jnp.mean(jnp.abs(factor) ** 2)
+    large_value_penalty /= len(decomposition)
+    total_loss += self.hypers.large_value_penalty_weight * large_value_penalty
+
+    return jnp.real(total_loss)
+
 
 def l2_loss_complex(x: jnp.ndarray, y: jnp.ndarray) -> jnp.ndarray:
   """Elementwise L2 loss for complex numbers."""
@@ -117,6 +255,18 @@
   return hyper.zipit([
-      hyper.uniform('init_scale', hyper.interval(0.2, 1.5)),
-      hyper.uniform('learning_rate', hyper.interval(0.05, 0.3)),
+      hyper.uniform('init_scale', hyper.interval(0.1, 1.0)),
+      hyper.uniform('learning_rate', hyper.interval(0.01, 0.2)),
+      hyper.uniform('discretization_weight', hyper.interval(0.0, 0.1)),
+      hyper.uniform('hallucination_prob', hyper.interval(0.0, 0.2)),
+      hyper.uniform('hallucination_scale', hyper.interval(0.0, 0.2)),
+      hyper.uniform('noise_std', hyper.interval(0.0, 0.01)),
+      hyper.uniform('target_noise_std', hyper.interval(0.0, 0.01)),
+      hyper.uniform('weight_decay', hyper.interval(0.00001, 0.001)),
+      hyper.uniform('clip_min', hyper.interval(0.0, 0.5)),
+      hyper.uniform('clip_max', hyper.interval(1.0, 3.0)),
+      hyper.uniform('large_value_penalty_weight', hyper.interval(0.0, 0.01)),
+      # Add noise to the gradient to aid in exploration.
+      hyper.uniform('grad_noise_std', hyper.interval(0.0, 0.001)),
+      hyper.uniform('half_int_start', hyper.interval(0.0, 1.0)),
   ])
 # EVOLVE-BLOCK-END
\end{lstlisting}
\end{minipage}
\end{adjustbox}
\caption{Magnified version of~\Cref{fig:relaxed-opt-diff}(left), giving the program that discovers a faster algorithm to multiply $4\times4$ matrices  (\emph{3/3}). Here \texttt{hyper} is a user-provided library for generating hyperparameter sweeps.}\label{fig:relaxed-opt-diff-appendix-3}
\centering
\end{figure}

\clearpage
\section{Details of mathematical discoveries of \method}\label{app:maths}

The data and verification code for all constructions reported in this section appear in \ResultsColab.

\subsection{First autocorrelation inequality}
For any function $f: \mathbb{R} \rightarrow \mathbb{R}$, define the \emph{autoconvolution} of $f$ as $$f*f (t) \coloneqq \int_\mathbb{R} f(t-x) f(x)\ dx.$$
Let $C_1$ denote the largest constant satisfying
\begin{equation}\label{maxf}
 \max_{-1/2 \leq t \leq 1/2} f*f(t) \geq C_1 \left(\int_{-1/4}^{1/4} f(x)\ dx\right)^2
\end{equation}
for all non-negative $f: \mathbb{R} \rightarrow \mathbb{R}$.  This problem arises in additive combinatorics, relating to the size of Sidon sets.  It is currently known that
$$1.28 \leq C_1 \leq 1.5098,$$
with the lower bound achieved in \cite{cloninger2017suprema} and the upper bound achieved in \cite{matolcsi2010improved} via a step function construction.
\method found a step function with 600 equally-spaced intervals on $[-1/4,1/4]$ that gives a slightly better upper bound $C_1 \leq 1.5053$.

\subsection{Second autocorrelation inequality}
Let $C_2$ be the smallest constant for which one has
$$ \|f*f\|_2^2 \leq C_2 \|f*f\|_1 \|f*f\|_\infty$$
for all non-negative $f:\mathbb{R} \rightarrow \mathbb{R}$.
It is known that
$$ 0.88922 \leq C_2 \leq 1$$
with the lower bound coming from a step function construction \cite{matolcsi2010improved}. 
\method found a step function with 50 equally-spaced intervals on $[-1/4,1/4]$ that gives a slightly better lower bound $0.8962 \leq C_2$.

\subsection{Third autocorrelation inequality}
Let $C_3$ be the largest constant satisfying
$$ \max_{-1/2 \leq t \leq 1/2} \left|f*f(t)\right| \geq C_3 \left(\int_{-1/4}^{1/4} f(x)\ dx\right)^2$$
for any function $f: \mathbb{R}  \rightarrow \mathbb{R}$. Clearly $C_3 \leq C_1$, since we now allow $f$ to take positive and negative values.  
There is a step function that gives the upper bound $C_3 \leq 1.45810$ \cite[page 75]{vinuesa2010generalized}.
\method found a step function with 400 equally-spaced intervals on $[-1/4,1/4]$ that gives a slightly better upper bound $C_3 \leq 1.4557$.

\subsection{An uncertainty inequality}

Given a function $f: \mathbb{R} \to \mathbb{R}$, define the Fourier transform $\hat f(\xi) := \int_\mathbb{R} f(x) e^{-2\pi i x\xi}\ dx$ and
$$ A(f) := \inf \{r > 0: f(x) \geq 0 \hbox{ for all } |x| \geq r \}.$$
Let $C_4$ be the largest constant for which one has
$$ A(f) A(\hat f) \geq C_4$$
for all even $f$ with $\max(f(0), \hat f(0)) < 0$.  It is known~\cite{gonccalves2017hermite} that
$$ 0.2025 \leq C_4 \leq 0.3523.$$
(The upper bound is stated as $0.353$ in the paper, but rounding their solution to the fourth digit gives $0.3523$).
We improved the upper bound to $C_4 \leq 0.3521$ with a similar linear combination as in~\cite{gonccalves2017hermite}, but with refined constants that were found by \method.

To obtain upper bounds for $C_4$, one constructs a specific "test function" $f$ satisfying the conditions and calculates the value $A(f)A(\hat f)$ for this function, which provides an upper bound $C_4 \le A(f)A(\hat f)$. Following the approach in~\cite{gonccalves2017hermite}, the test function is sought in the form $f(x) = P(x)e^{-\pi x^2}$, where $P(x)$ is an even polynomial constructed as a linear combination of Hermite polynomials $H_{4k}(x)$. This form is particularly useful because the Fourier transform of $H_n(x)e^{-\pi x^2}$ is $i^n H_n(\xi)e^{-\pi \xi^2}$. For an even polynomial $P(x) = \sum c_{4k} H_{4k}(x)$, the Fourier transform of $f(x)$ is $\hat f(\xi) = \sum c_{4k} i^{4k} H_{4k}(\xi)e^{-\pi \xi^2} = (\sum c_{4k} H_{4k}(\xi))e^{-\pi \xi^2} = P(\xi)e^{-\pi \xi^2}$. Thus, $A(f)$ is related to the largest positive root of $P(x)$, and $A(\hat f)$ is related to the largest positive root of $P(\xi)$. Specifically, if $P(x) \ge 0$ for large $|x|$, $A(f)$ is the largest positive root of $P(x)$, and $A(\hat f)$ is the largest positive root of $P(\xi)$, implying $A(f) = A(\hat f)$. The inequality becomes $C_4 \le (A(f))^2$.

The method involves finding coefficients $c_0, c_1, c_2, \dots$ for the polynomial $P(x) = c_0 H_0(x) + c_1 H_4(x) + c_2 H_8(x) + \dots$ such that $P(x)$ satisfies certain constraints (related to $f(0)<0, \hat f(0)<0$ and being positive for large $|x|$) and minimizes the largest positive root of $P(x)$. In our approach, the polynomial $P(x)$ is constructed such that $P(0)=0$ (a condition used in the optimization process to simplify constraints), meaning $P(x)$ has a factor of $x^2$. The largest positive root $r_{\max}$ of $P(x)$ is then the largest positive root of $P(x)/x^2$. The upper bound on $C_4$ derived from this construction is $r_{\max}^2 / (2\pi)$.

The refined constants found by \method for $P(x) = c_0 H_0(x) + c_1 H_4(x) + c_2 H_8(x)$ are $[c_0, c_1, c_2] \approx [0.32925, -0.01159, -8.9216 \times 10^{-5}]$. Using these coefficients to construct $P(x)$, finding its largest positive root $r_{\max}$ (by finding the largest positive root of $P(x)/x^2$), and calculating $r_{\max}^2 / (2\pi)$ yields the improved upper bound $C_4 \leq 0.3521$. Qualitatively our linear combination is very similar to the one found in~\cite{gonccalves2017hermite}, thus empirically confirming their hypothesis the construction is nearly optimal.

\emph{Note}: After publishing the first version of this manuscript, Henry Cohn pointed out that in a recent paper~\cite{cohn2019optimal} they used a similar, but more refined approach to get the better constant 0.3284. By incorporating their refined approach into \method, we improved our reported constant further to 0.3216. For details, we refer to \ResultsColab.

\subsection{Erd\H{o}s' minimum overlap problem}
Let $C_5$ be the largest constant for which
$$ \sup_{x \in [-2,2]} \int_{-1}^1 f(t) g(x+t)\ dt\geq C_5$$
for all non-negative $f,g: [-1,1] \to [0,1]$ with $f+g=1$ on $[-1,1]$ and $\int f = 1$, where we extend $f,g$ by zero outside of $[-1,1]$. This constant controls the asymptotics of the Minimum Overlap Problem of \cite{erdHos1955some}.  The bounds
$$ 0.379005 \leq C_5 \leq 0.380927$$
are known, where the lower bound was obtained in \cite{white2023new} via convex programming methods.

\begin{figure}
    \centering
    \includegraphics[scale=0.5]{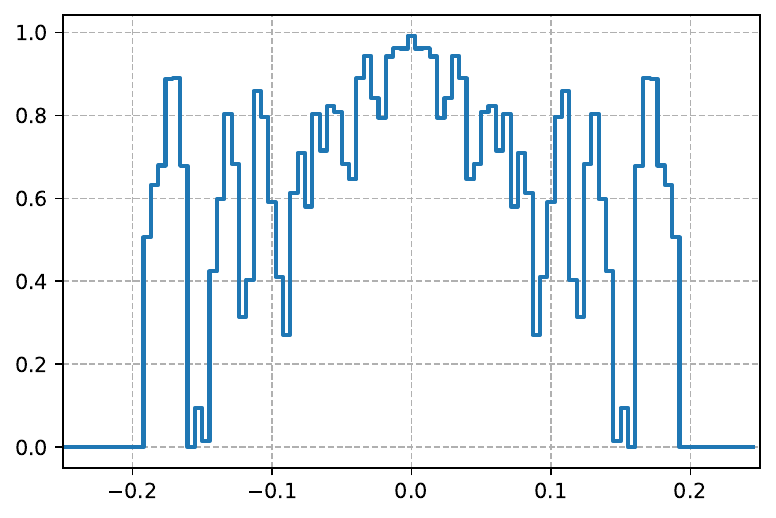}
    \caption{Construction found by \method for the minimum overlap problem of Erd\H{o}s.}
    \label{fig:erdos}
\end{figure}

It is known (see~\cite{haugland2016minimum}) that
this constant is equal to the infimum, over all step
functions $h$ on $[0, 2]$ with values in $[0, 1]$ and satisfying
$
\int_0^2 h(x)dx = 1
$
of
$$
\max_k \int h(x)(1 - h(x+k))dx 
.$$ The upper bound to the Erd\H{o}s minimum overlap problem was then obtained by using this result, in~\cite{haugland2016minimum} by a step function
construction. The step function depicted in \Cref{fig:erdos} does ever so slightly better than the previous bound, giving the upper bound of $C_5 \leq 0.380924$.

~

\subsection{Sums and differences of finite sets}
Let $C_6$ be the largest constant for which the following statement holds: there exist arbitrarily large finite sets of integers $A,B$ with $|A+B| \ll |A|$ and $|A-B| \gg |A+B|^{C_6}$.
(Here $A+B = \{a+b : a \in A, b \in B\}$ and $A-B = \{a-b : a \in A, b \in B\}$ denote the sumset and difference set, respectively. The notation $X \ll Y$ means that $X \le C Y$ for some constant $C$ independent of the sets $A,B$ (for sufficiently large sets $A,B$). The notation $X \gg Y$ means that $X \ge C' Y$ for some positive constant $C'$ independent of the sets $A,B$ (for sufficiently large sets $A,B$).)
\begin{equation}\label{cds}
 1.14465 \leq C_6 \leq \frac{4}{3};
\end{equation}
see \cite[Corollary 3]{gyarmati2007sums} for the 
upper bound and \cite[Theorem 1]{gyarmati2007sums} for the lower bound. The main tool for the lower bound is the following result of \citet{gyarmati2007sums}:
\begin{equation}\label{cds-lower}
 C_6 \geq 1 + \frac{\log \frac{|U-U|}{|U+U|}}{\log (2 \max (U) + 1)}
\end{equation}
for any finite set $U$ of non-negative integers containing zero satisfying $|U-U| \leq 2 \max (U) + 1$.
\method found a set $U_1$ of size 2003 improving the lower bound to $1.1479 \leq C_6$,
and another set $U_2$ of size 54265 further improving the lower bound  to 
$1.1584 \leq C_6$.

\subsection{Packing unit regular hexagons inside a regular hexagon}
\begin{figure}
    \centering
    \includegraphics[width=0.4\linewidth]{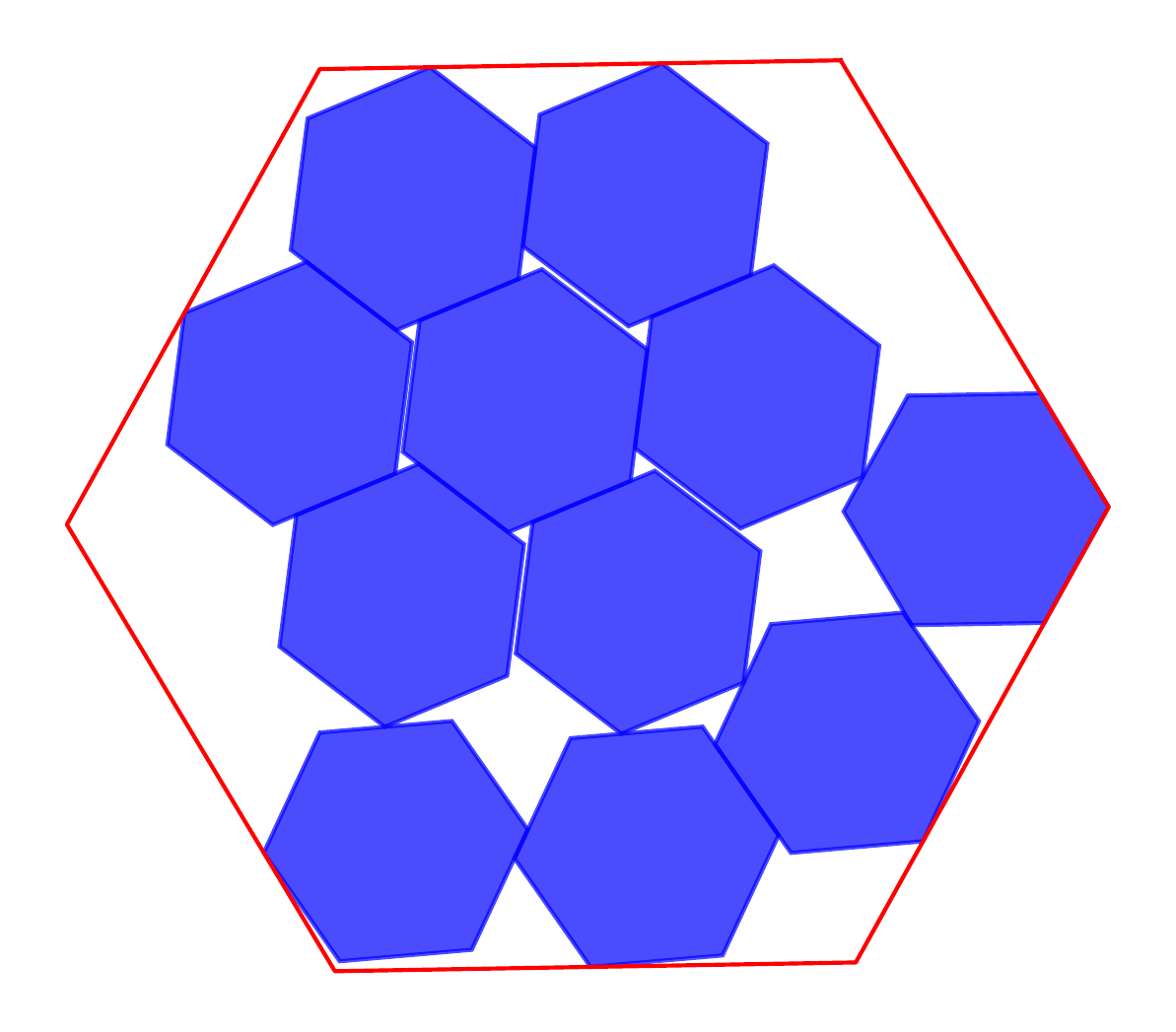}
    \hfill
    \includegraphics[width=0.4\linewidth]{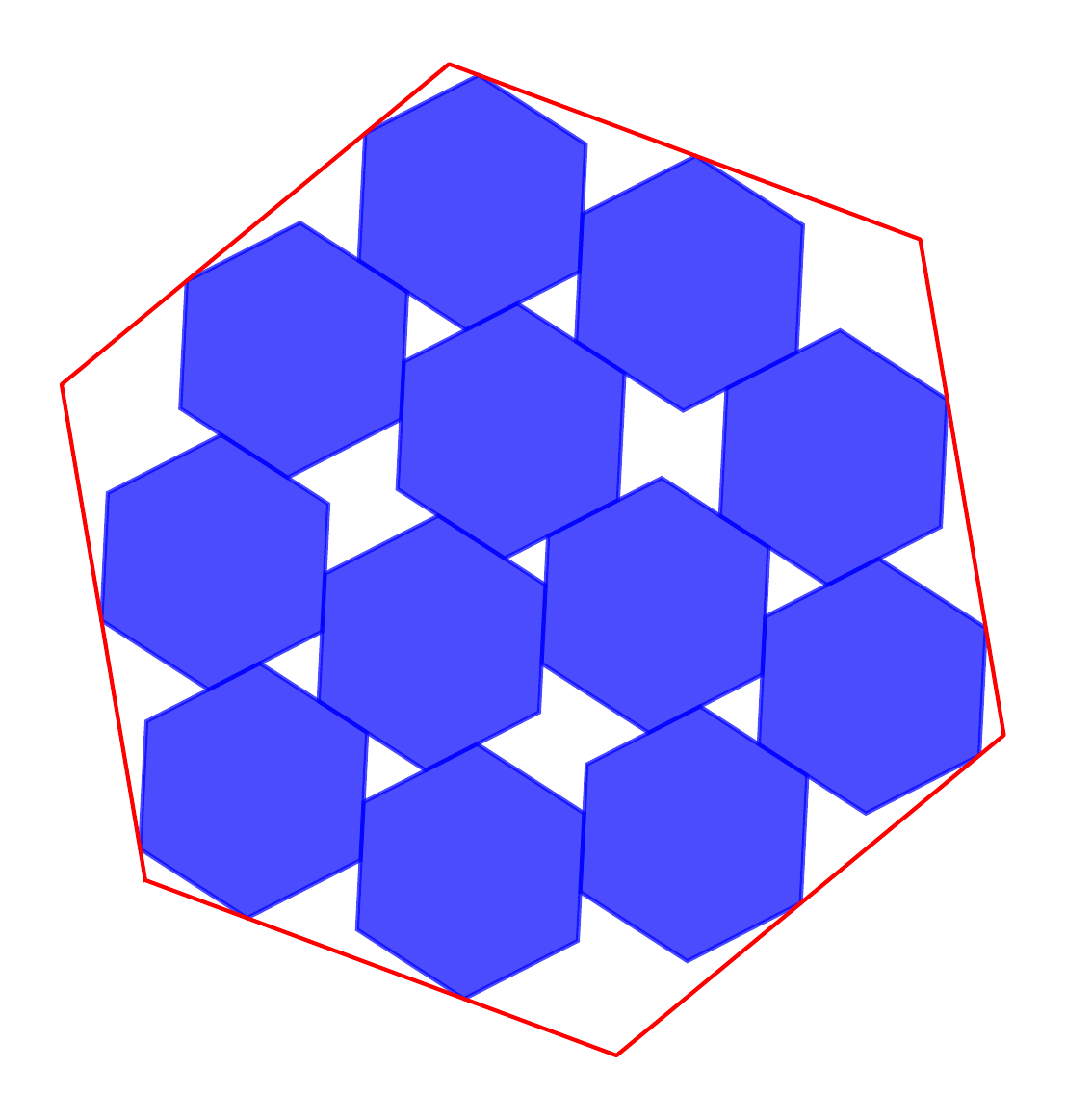}
    \caption{Constructions of the packing problems found by \method. Left: Packing 11 unit hexagons into a regular hexagon of side length 3.931. Right: Packing 12 unit hexagons into a regular hexagon of side length 3.942.
    \label{fig:hexagons}}
\end{figure}

Consider the problem of packing $n$ disjoint regular hexagons with unit side length into a larger regular hexagon, minimizing the side length of the outer hexagon.
For $n=11$ and $n=12$, the best known constructions use outer hexagons of side lengths $3.943$ and $4.0$, respectively \cite{geometry_collection}.
\method found packing arrangements
that improve these bounds to $3.931$ and $3.942$, respectively.
These arrangements are shown in \Cref{fig:hexagons}.

\subsection{Minimizing the ratio of maximum to minimum distance}
\begin{figure}
    \centering
    \includegraphics[width=0.45\linewidth]{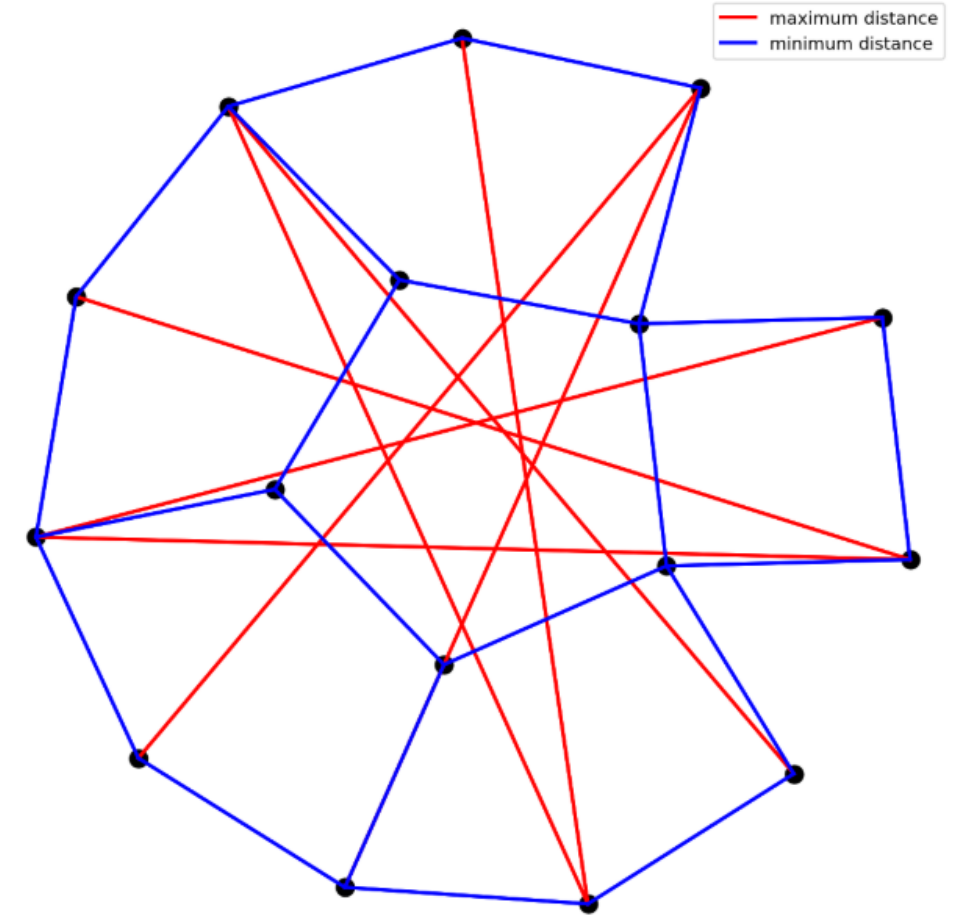}
    \includegraphics[width=0.45\linewidth]{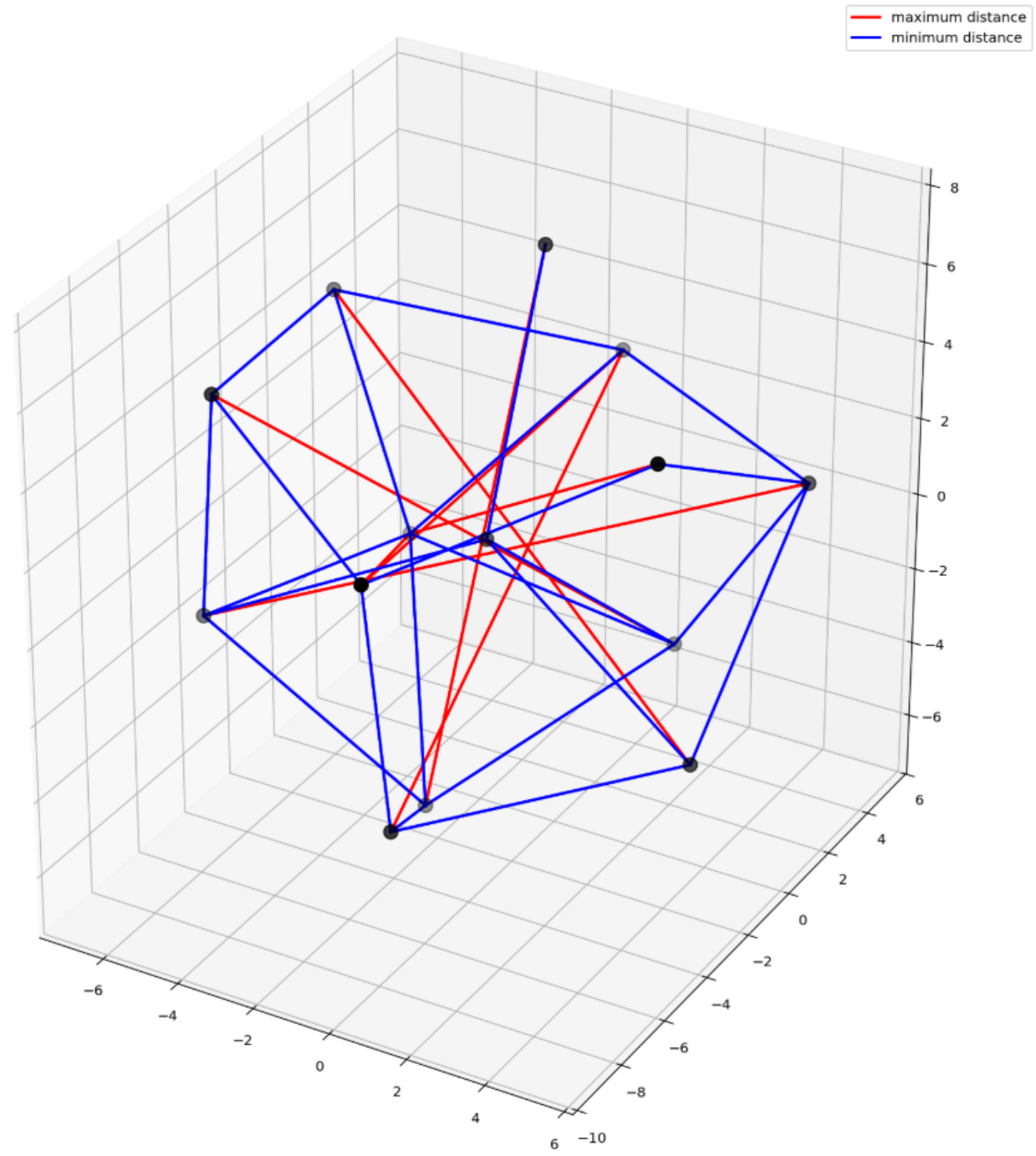}
    \caption{Left: 16 points in 2 dimensions achieving a ratio of maximum distance to minimum distance of
    $\approx \sqrt{12.889266112}$.     
    Right: 14 points in 3 dimensions achieving a ratio of $\approx \sqrt{4.165849767}$. Both constructions improve the best known bounds.}
    \label{fig:distance_ratios}
\end{figure}

For any $n$ and $d$, the goal of this problem is to find $n$ points in the $d$-dimensional space so as to minimize the ratio between the maximum and minimum pairwise distances. \method found two new constructions improving the best known bounds.
The found constructions are shown in \Cref{fig:distance_ratios}.

In 2 dimensions, \method found 16 points with ratio $\approx \sqrt{12.889266112}$, improving the best known bound of $\sqrt{12.890}$~\citep{geometry_collection}.
(In this reference, instead of the ratio itself, the square of the ratio is reported, and we use the same convention.)
 
In 3 dimensions, \method found 14 points with ratio $\approx \sqrt{4.165849767}$, improving the best known bound of $\sqrt{4.168}$~\citep{geometry_collection}.

\subsection{The Heilbronn problem for triangles}
The goal of this problem is to find $n$ points on or inside a triangle with unit area so that the area of the smallest triangle formed by these points is maximized. For $n=11$, the SOTA was $0.036$~\citep{geometry_collection}, and AlphaEvolve found a construction with minimum area larger than $0.0365$, which is shown in \Cref{fig:heilbronn} (left). 

\subsection{The Heilbronn problem for convex regions}
The goal of this problem is to find $n$ points on or inside a convex region with unit area so that the area of the smallest triangle formed by these points is maximized. \method improved two of the best known bounds.

\begin{figure}
    \centering
    \includegraphics[width=0.25\linewidth]{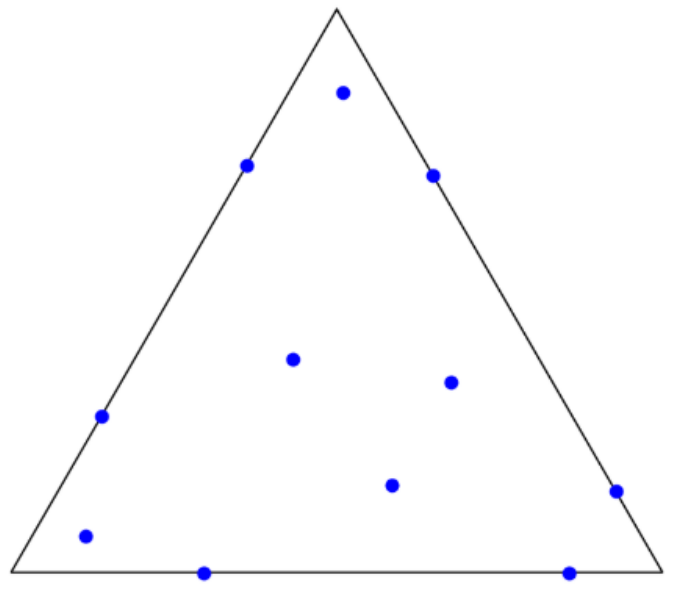}
    \includegraphics[width=0.25\linewidth]{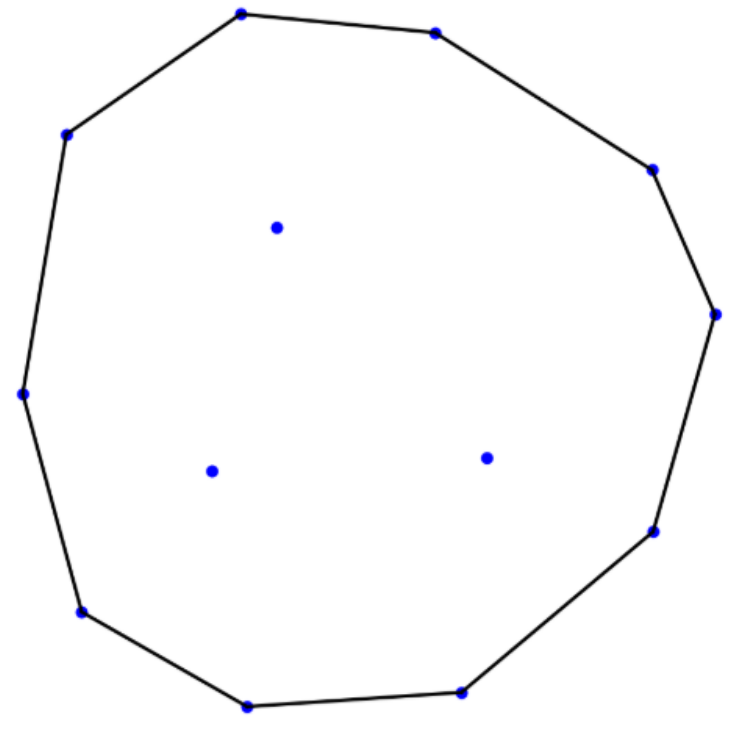}
    \includegraphics[width=0.25\linewidth]{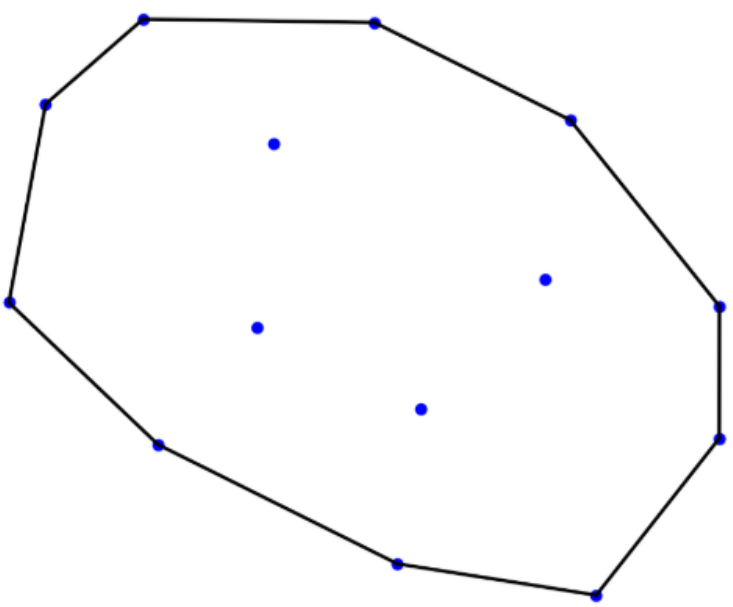}
    \caption{New constructions found by \method improving the best known bounds on two variants of the Heilbronn problem. Left: 11 points in a unit-area triangle with all formed triangles having area $\geq 0.0365$. Middle: 13 points inside a convex region with unit area with all formed triangles having area $\geq 0.0309$. Right: 14 points inside a unit convex region with minimum area  $\ge 0.0278$.}
    \label{fig:heilbronn}
\end{figure}

For $n=13$, the SOTA was $0.0306$~\citep{geometry_collection}, and \method improved it to $0.0309$ (see \Cref{fig:heilbronn} (middle)).
For $n=14$, the SOTA was $0.0277$~\citep{geometry_collection} and \method improved it to $0.0278$ (see \Cref{fig:heilbronn} (right)).

\subsection{Kissing number in dimension 11}
The kissing problem asks how many disjoint unit spheres can be packed tangent to a given unit sphere. The maximum such number in $d$ dimensions is called the \emph{$d$-dimensional kissing number~\citep{kissing_survey}}.
For $d=11$, the best known lower bound was 592~\citep{kissing_11} and \method improved this to 593. To prove the lower bound of 593 for the kissing number in dimension 11, \method found 593 many 11-dimensional non-zero points with integral coordinates such that  
the maximum norm of these points is smaller than their minimum pairwise distance.
By the following lemma, this implies the kissing number in dimension 11 is at least 593.

\begin{lemma}
Let $C \subset \mathbb{R}^d$ be a set of points satisfying
$0 \notin C$ and 
\begin{equation*}
    \min \left\{ \|x-y\| : x \neq y \in C \right\} \ge \max \left\{ \|x\|: x \in C \right\}.
\end{equation*}
Then unit spheres centred at $\left\{  \frac{2x}{\|x\|}:{x\in C}\right\}$ form a valid kissing configuration in dimension $d$. In particular, the kissing number in dimension $d$ is at least $|C|$.
\end{lemma}

\begin{proof}
For any $x\neq y \in C$, the inequality $\|x - y\|^2 \geq \max \{\|x\|^2, \|y\|^2\}$ implies
\begin{equation}
2 \langle x, y \rangle \leq \|x\|^2 + \|y\|^2 - \max \{\|x\|^2, \|y\|^2\} = \min \{\|x\|^2, \|y\|^2\} \leq \|x\|\cdot\|y\|,
\label{eq:abi}
\end{equation}
where the last inequality holds because the minimum of two positive numbers is less than or equal to their geometric mean.
The points $\left\{ \frac{2x}{\|x\|}:{x\in C}\right\}$ have norm 2, so unit spheres centred at them are tangent to a unit sphere centred at the origin.
The last step is to show that these spheres do not overlap. This is equivalent to showing, for all $x\neq y \in C$, that
\[
\left\| \frac{2x}{\|x\|} -  \frac{2y}{\|y\|}\right\| \geq 2.
\]
After simplifying, this is equivalent to
\(
2 \langle x, y \rangle  \leq \|x\|\cdot\|y\|\),
which we have proved in \eqref{eq:abi}.
Thus unit spheres centred at $\left\{ \frac{2x}{\|x\|}:{x\in C}\right\}$ form a valid kissing configuration in dimension $d$, as required.
\end{proof}

\subsection{Packing circles inside a unit square to maximize sum of radii}

Given a positive integer $n$, the problem is to pack $n$ disjoint circles inside a unit square so as to maximize the sum of their radii. \method found two new constructions improving the state of the art~\citep{geometry_collection}.

For $n=26$, the SOTA was $2.634$, and AlphaEvolve improved it to $2.635$; see \Cref{fig:circle_packing} (left).
For $n=32$, the SOTA was $2.936$, and AlphaEvolve improved it to $2.937$; see \Cref{fig:circle_packing} (middle).

\subsection{Packing  circles inside a rectangle of perimeter 4 to maximize sum of radii}
\begin{figure}[b]
    \centering
    \includegraphics[width=0.25\linewidth]{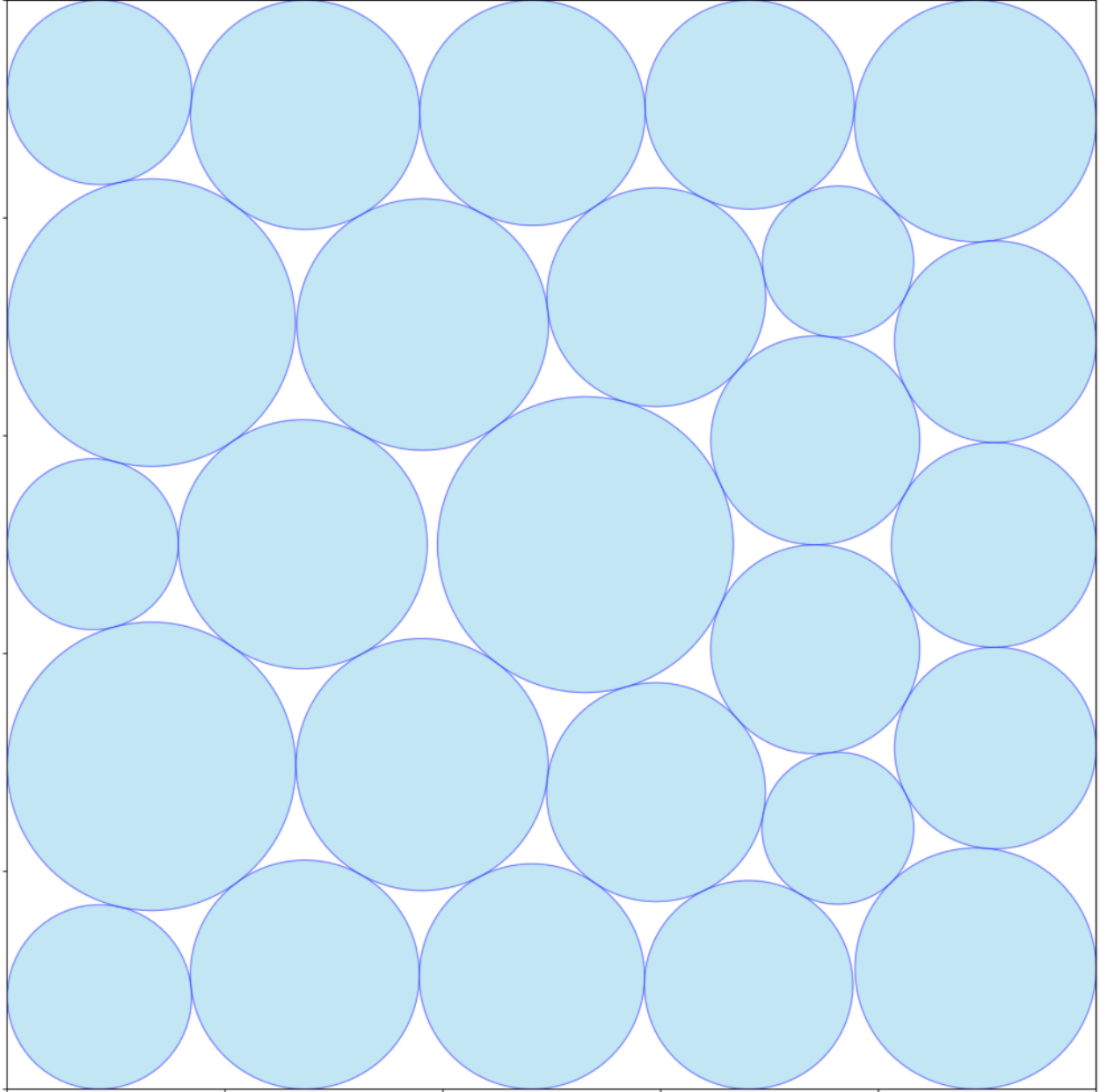}
    \includegraphics[width=0.25\linewidth]{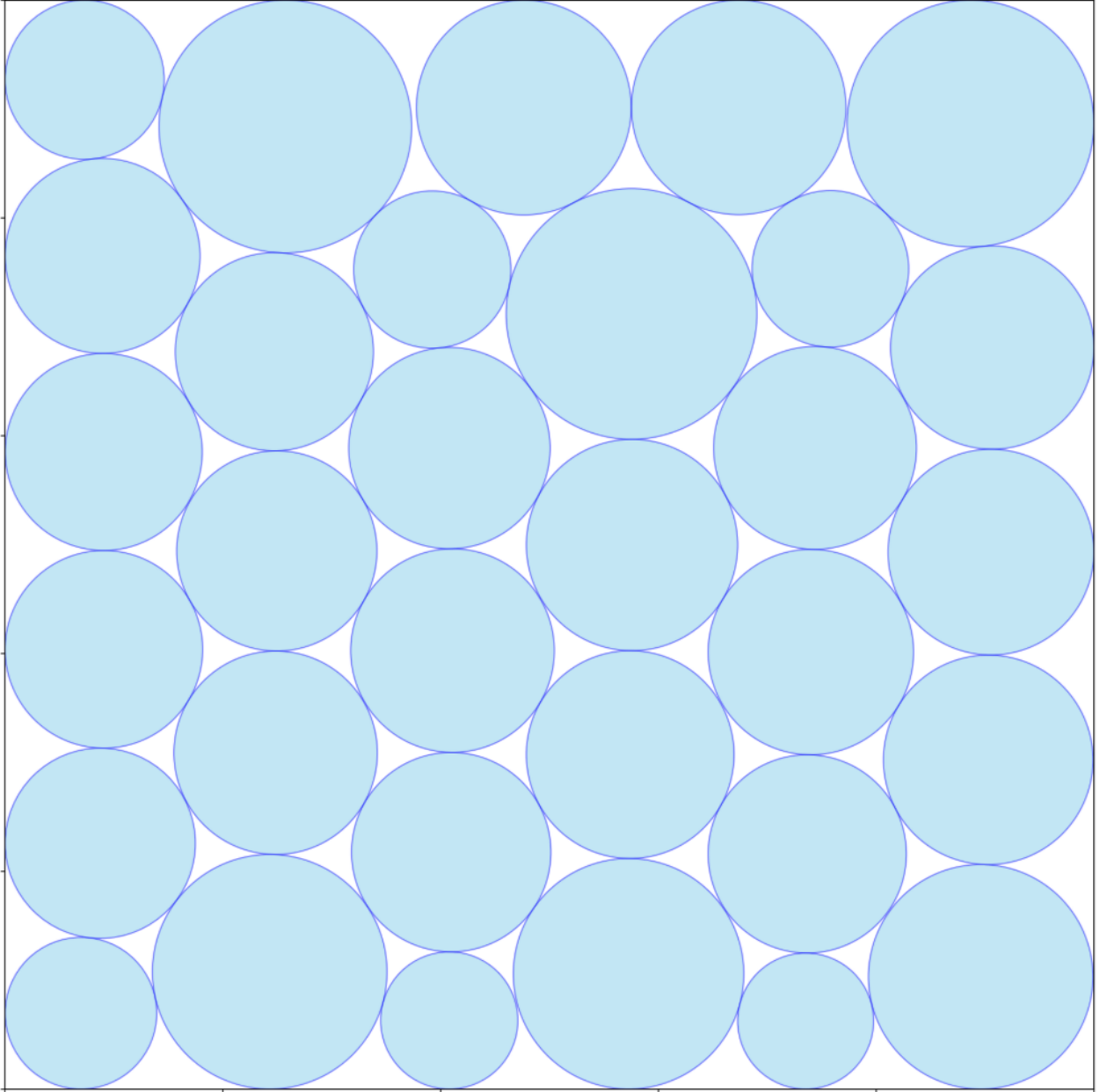}
    \includegraphics[width=0.25\linewidth]{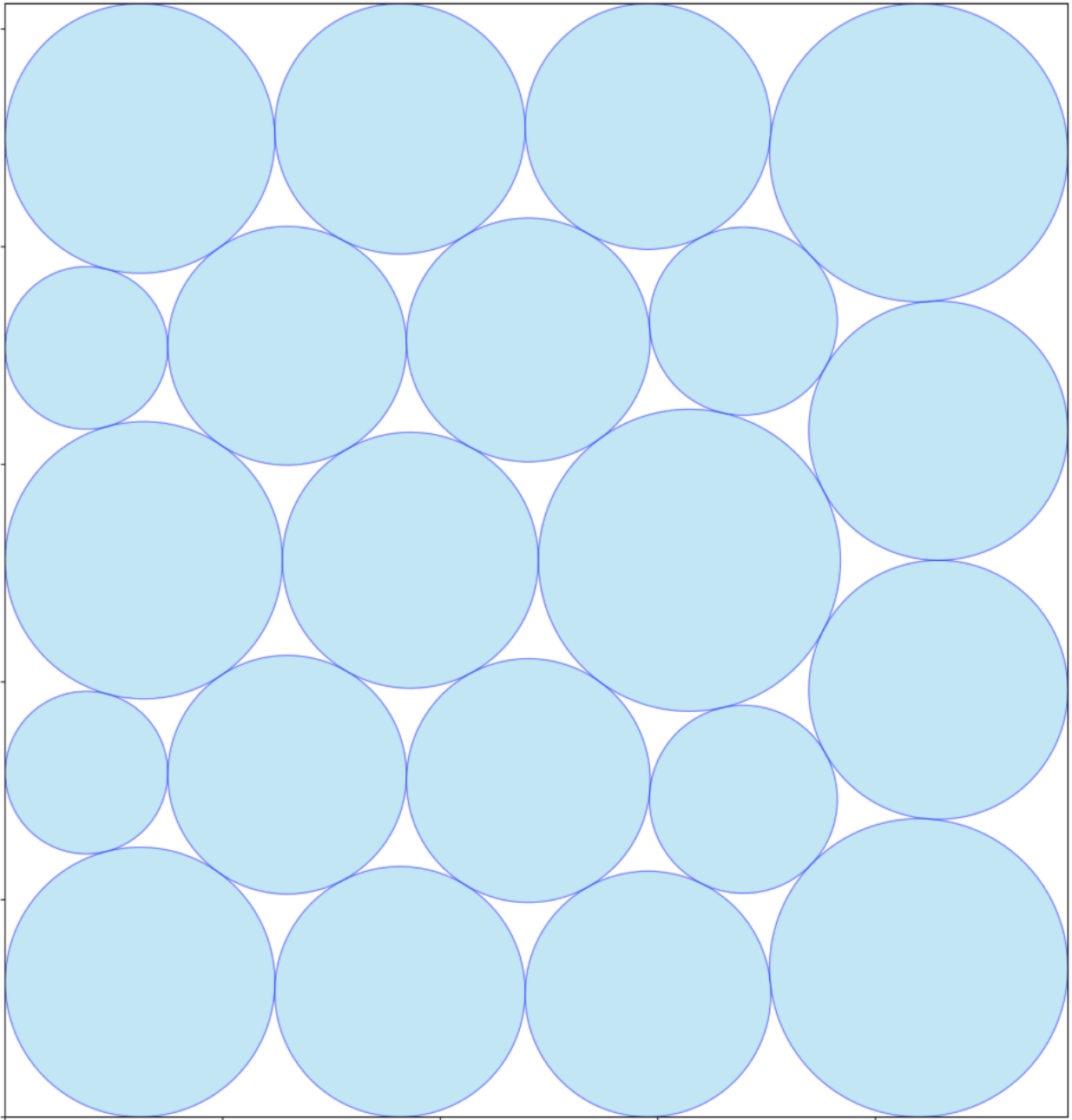}
    \caption{New constructions found by \method improving the best known bounds on packing circles to maximize their sum of radii. Left: $26$ circles in a unit square with sum of radii $\geq 2.635$. Middle: $32$ circles in a unit square with sum of radii $\geq 2.937$. Right: 21 circles in a rectangle with perimeter $4$, with sum of radii $\geq 2.365$.}
    \label{fig:circle_packing}
\end{figure}

Given a positive integer $n$, the problem is to pack $n$ disjoint circles inside a rectangle of perimeter 4 so as to maximize the sum of their radii. 
\method found a new construction for $n=21$, improving the state of the art from $2.364$~\citep{geometry_collection} to $2.3658$; see \Cref{fig:circle_packing} (right).

\end{document}